\documentclass[11pt]{article}

\usepackage[preprint]{acl}

\usepackage{times}
\usepackage{amsmath}   % equation, aligned, \text, \mathbb, \mathrm, \!, \;, \phantom
\usepackage{amssymb}   % \mathbb (if not already from amsmath/amsfonts)
\usepackage{latexsym}
\usepackage{booktabs}
\usepackage{multirow}
\usepackage{soul,color,xcolor}
\usepackage{booktabs}
\usepackage{tabularx}
\usepackage{makecell}
\usepackage{mdframed}

\newcommand{\namess}{\textsc{SHALA-LLM}} %sans space
\newcommand{\names}{\textsc{SHALA-LLM }}
\newcommand{\namesm}{\textsc{Smartly Handling Ambiguous Labels in Aligning LLMs}}

\newcommand{\namesr}{\textsc{SHALA}}

\newcommand{\jw}[1]{\sethlcolor{cyan}\hl{[Jingyao: #1]}}
\newcommand{\paul}[1]{\textcolor{red}{paul: #1}}
\usepackage[T1]{fontenc}
\usepackage{comment}

\usepackage[utf8]{inputenc}

\usepackage{microtype}

\usepackage{inconsolata}

\usepackage{graphicx}

\title{\namess: Smartly Handling Ambiguous Labels in Aligning LLMs} 

\author{
Jingyao Wu$^{1}$\thanks{Equal contribution.},
Ashley Wang$^{1}$\footnotemark[1],
Keane Ong$^{1,2}$,
Paul Pu Liang$^{1}$,
Rosalind W. Picard$^{1}$ \\$^{1}$MIT Media Lab, Massachusetts Institute of Technology \\
$^{2}$National University of Singapore \\
\texttt{\{jingyaow,ashley25,keaneong,ppliang\}@mit.edu}\\
\texttt{picard@media.mit.edu}
}
\begin{document}
\maketitle
\begin{abstract}
Many human-centered tasks, including natural language inference (NLI) and emotion recognition (ER), 
%r are inherently subjective and ambiguous, 
have multiple plausible interpretations, leading to label ambiguity and challenging disagreements across human annotators. %\paul{state and motivate the problem for training llms?}. 
%\keane{Human-centered tasks, such as natural language inference and emotion recognition, are inherently subjective, often leading to annotator disagreement on the correct label. As LLMs are increasingly deployed in real-world settings, representing this ambiguity is essential for exposing which inputs are contested, so downstream users can distinguish clear cases from genuinely ambiguous ones. Yet, LLM alignment has predominantly centered around the assumption that each input has one correct label, often excluding annotator disagreement from the training process. To address this, we introduce SHALA-LLM..}
As LLMs are increasingly deployed in real-world settings, faithfully modeling such ambiguity is essential to identify contested inputs, preserve variability in ambiguous cases, and capture the full distribution of human judgments. Yet, existing LLM alignment approaches have predominantly assumed a single correct label, excluding annotator disagreement during optimization. Instead of treating this ambiguity as noise, we show how to treat it as information that improves model behavior through a new algorithm called \namesm~(\namess). This reinforcement learning framework provides a new way for
%r introducing Smart  this work, we introduce \names: \namesm %\textit{AmbED: Ambiguity-Enhanced Distributional Supervision}, a reinforcement learning based alignment framework that enables 
LLMs to learn directly from annotator distributions while dynamically prioritizing highly ambiguous samples during optimization.
%Experiments on ambiguity-sensitive NLI and ER benchmarks, including ChaosNLI, GoEmotions, and MSP-Podcast, demonstrate that \names consistently improves distributional alignment \paul{define what distributional alignment means + motivate its importance} and conventional classification performance compared with conventional single-label supervision. For example, on ChaosNLI, \names reduces Jensen-Shannon Divergence by 62.1\% while improving weighted F1 by 12.1\% compared with majority-label supervision. 
Experiments on ambiguity-sensitive NLI and ER benchmarks, including ChaosNLI, GoEmotions, and MSP-Podcast, demonstrate that \names\ improves agreement with annotator label distributions, e.g. on ChaosNLI, it reduces Jensen–Shannon Distance by up to 62.1\%. At the same time, \names\ improves F1 by up to 16.7\%, showing that modeling annotator disagreement can also strengthen classification performance\footnote{Code will be available upon publication}.

\end{abstract}

\section{Introduction}
In many human-centered tasks, different individuals may interpret the same text, speech, or interaction differently depending on contextual understanding, personal experience, cultural background, or emotional perception~\cite{uma2021learning}. This phenomenon is especially common in natural language inference (NLI)~\cite{nie-etal-2020-learn, chen-etal-2025-rose} and emotion recognition (ER)~\cite{sethu2019ambiguous, wu2026amber} %rwhere multiple plausible interpretations coexist, 
leading to disagreement across human annotators and complicating the alignment of large language models (LLMs).
%rthat contains nuanced and informative signals regarding uncertainty, interpretational diversity, and human perception.
Most existing LLM alignment paradigms treat disagreement as annotation noise, collapsing ambiguous annotations into single target labels through majority voting, label averaging, or calibration methods that assume a single gold standard~\cite{radharapu2025arbiters}. In doing so, they discard the disagreement embedded within human annotations, particularly in highly ambiguous settings where annotator uncertainty and conflicting judgments are prominent~\cite{baan2022stop}. %\paul{there is some work in distributional alignment of LLMs eg \url{https://arxiv.org/abs/2411.05403}. also pluralistic alignment \url{https://arxiv.org/abs/2402.05070}. we either explain these fields exist, say we propose a method and compare to them. or explain why your alignment problem is different from them (while still being important)  } \jw{Updated as below.} 
Although recent studies have explored distributional and pluralistic alignment in LLMs, including uncertainty-aware distribution elicitation and alignment with diverse human viewpoints or preferences~\cite{sorensen2024roadmap, meister2025benchmarking}, they primarily operate at inference time through prompting or distribution estimation, rather than directly optimizing LLMs with annotator disagreement during training, limiting learning dynamics and model adaptation under highly ambiguous settings~\cite{baan2022stop}.

% We present a novel solution \namess: 
% \namesm, \paul{explain in 1 sentence how the method works + the key idea. eg. SHALA-LLM dynamically reweights model predictions according to the degree of annotator disagreement, enabling highly ambiguous samples to exert greater influence during learning}
% providing reinforcement learning alignment that directly aligns LLMs with distributions of human judgments under ambiguous supervision 
% %r Unlike existing paradigms that collapse disagreement into dominant labels, \names enables ambiguity-aware alignment directly from human disagreement under subjective supervision. 
% %r Further, \names aligns LLM predictions %r with distributions of human judgments 
% while \textcolor{blue}{dynamically reweights model predictions according to the degree of annotator disagreement}. This novel method allows disagreement structures to exert greater influence during policy optimization.

In this paper, we present \namess: \namesm, a new ambiguity-aware distributional alignment framework that directly learns from human disagreement under ambiguous supervision. \namess\ dynamically reweights rollout rewards according to the degree of annotator disagreement, enabling highly ambiguous samples containing richer disagreement structures to exert greater influence during learning while remaining aligned with the underlying distributions of human judgments under ambiguous supervision.

%Consequently, highly ambiguous samples exert greater influence during policy optimization, enabling ambiguity-aware learning directly from human disagreement structures.

Evaluations on ambiguity-sensitive NLI and ER benchmarks, including ChaosNLI \cite{nie-etal-2020-learn}, GoEmotions \cite{2020-goemotions}, and MSP-Podcast \cite{msp-podcast}, demonstrate that \names consistently improves distributional alignment and classification results compared with majority-label supervision. On ChaosNLI, \names reduces Jensen-Shannon Distance (JSD) by 62.1\% while improving F1 by 16.6\%; and on MSP-Podcast, \names reduces JSD by 6.2\% and improving F1 by 29.2\%. Systematic analyses reveal \names exhibits stronger robustness under highly ambiguous conditions, provides particularly strong benefits for inherently ambiguous semantic categories, and encourages reasoning behaviors that better reflect uncertainty and multiple plausible interpretations under ambiguous supervision. This demonstrates that preserving disagreement structures leads to more robust and human-aligned LLM behavior under ambiguity.

\section{Related Work}
%\jw{@ashely, can you put whatever literature review you did in your thesis into the corresdponding sections here?}

%\paragraph{Annotator Ambiguity in NLP Datasets.} Annotations for human-centered tasks, often arise from multiple human annotators that may disagree. This disagreement may not inherently represent annotation noise, but rather an informative signal reflecting uncertainty, subjectivity, and diverse human interpretations \cite{aroyo2015truth, davani2022dealing, chen2024seeing, plank-2022-problem}. This phenomenon has been studied across multiple domains, including semantic reasoning tasks such as natural language inference (NLI)~\cite{jayaweera2025disagreement, pavlick-kwiatkowski-2019-inherent} and subjective perception tasks such as emotion recognition and affective computing ~\cite{aer-llm, chou-2025, wu2024must, sethu2019ambiguous}. Such ambiguity may manifest as disagreement over categorical labels (e.g., entailment classes or discrete emotions) or variability in continuous perceptual dimensions such as arousal and valence. Yet, to our knowledge, modelling approaches largely do not account for ambiguous signals.

% However, most existing learning frameworks assume the existence of a single ``gold-standard'' ground-truth label and therefore simplify disagreement through majority voting or label averaging. While such formulations are convenient for conventional supervised learning, they often discard potentially informative disagreement structures and ambiguity in human interpretations.
\paragraph{Modeling ambiguity in NLP.} Annotations for human-centered tasks from multiple human annotators may disagree, reflecting important information
\cite{aroyo2015truth, davani2022dealing, chen2024seeing, plank-2022-problem}. This phenomenon has been studied across multiple domains, including semantic reasoning tasks such as natural language inference (NLI)~\cite{jayaweera2025disagreement, pavlick-kwiatkowski-2019-inherent} and subjective perception tasks such as emotion recognition and affective computing ~\cite{niu2025rethinking, chou-2025, wu2024must, sethu2019ambiguous}. %Such ambiguity may manifest as disagreement over categorical labels (e.g., entailment classes or discrete emotions) or variability in continuous perceptual dimensions such as arousal and valence. 
Researchers increasingly recognize the importance of addressing annotation ambiguity %r, ambiguity-aware modelling approaches remain relatively underexplored. %r in NLP and affective computing. 
%rAmong the limited existing studies, 
with prior work in categorical representation settings such as NLI and discrete ER investigating soft-label supervision~\cite{wu2026amber, fard2025affectnet}, multi-label formulations~\cite{ando19_interspeech, ju2020}, and disagreement-aware learning approaches~\cite{chou2019every}. In continuous affect prediction settings, others have explored %r distributional representations including
Gaussian Distribution~\cite{gaussian}, Beta Distribution~\cite{beta, wu2024dual}, and non-parametric label distributions~\cite{Wu2022ANS}. However, these approaches have been primarily developed for conventional supervised learning, while LLM alignment needs advancing.  
\begin{figure*}[!tb]
\centering
\includegraphics[width=\linewidth, height = 6.7cm]{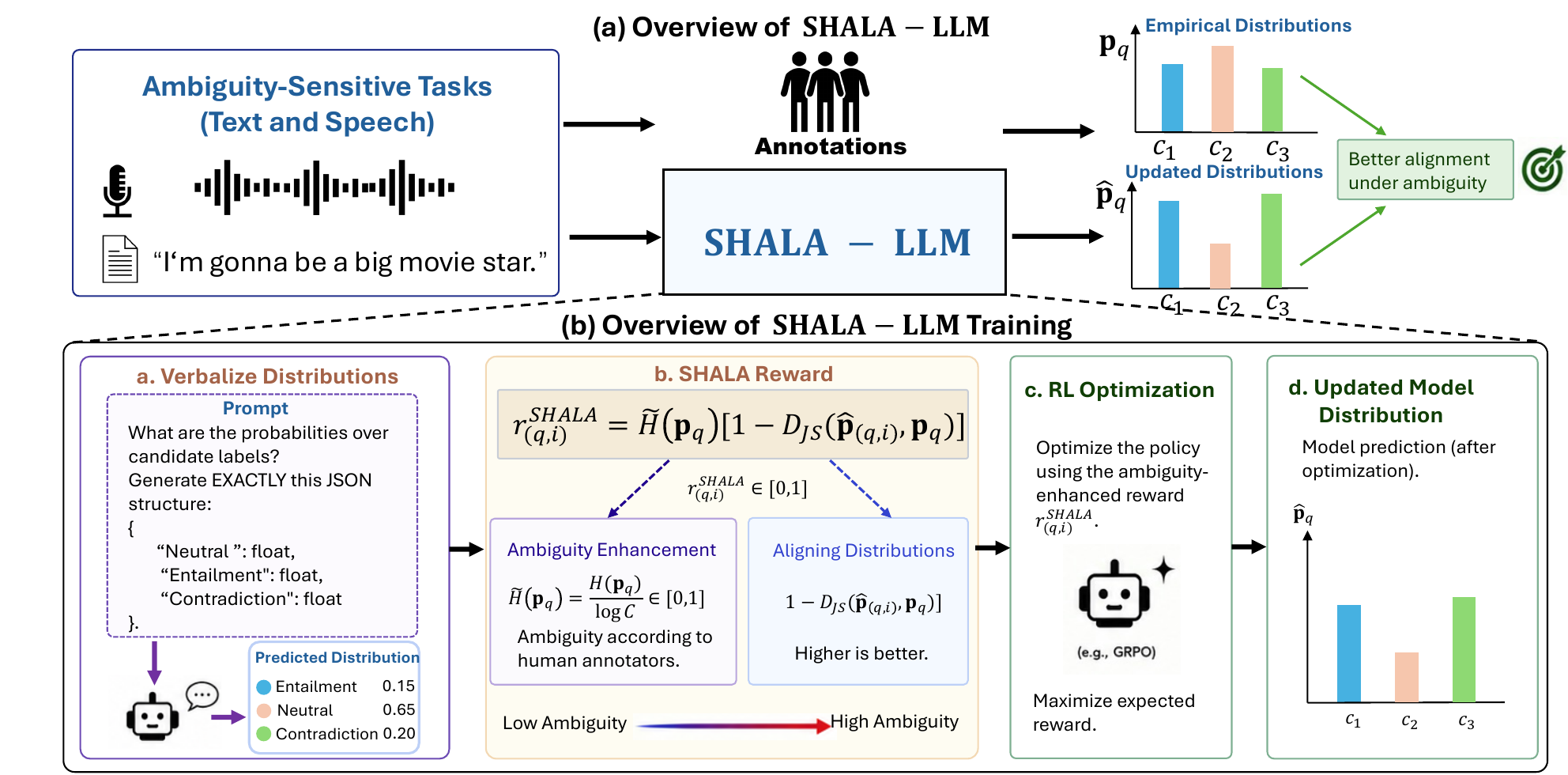}
\caption{Overview of proposed \namess: \namesm. %\paul{write a full standalone caption for the method} 
Given ambiguity-sensitive tasks such as NLI and ER (a), annotations from multiple annotators are aggregated into empirical label distributions. \namess~ prompts the LLM to verbalize probability distributions over candidate classes (b-a), which are directly aligned with annotator distributions through \namesr~ reward (b-b). Rollout rewards are dynamically reweighted according to annotator disagreement during GRPO optimization (b-c), producing distributions that better capture uncertainty and diverse human interpretations under ambiguous supervision (b-d).} %\paul{too many parts to the figure. separate into an overview of the problem/annotations/distribution + separate figure for the framework} }
\label{fig:ADS}
\end{figure*}%\vspace{-1em}

\paragraph{LLM alignment for ambiguous labels.}
As LLMs expand use in human-centered applications, they increasingly encounter ambiguous labels. To this end, recent studies have begun analyzing how LLMs behave under annotator disagreement and subjective supervision settings~\cite{lu2025llm, jia2026decoding}. Efforts have explored ambiguity-aware instruction fine-tuning~\cite{aer-llm}, soft-label supervision~\cite{davani-etal-2022-dealing, chen-etal-2025-rose, chen-etal-2024-seeing}, distributional alignment~\cite{meister2025benchmarking} and disagreement-aware alignment strategies~\cite{baan2022stop} to better align LLM outputs with distributions of human judgments. Smart handling of ambiguous labels for alignment remains to be extended to RL reasoning methods such as Group Relative Policy Optimization (GRPO)~\cite{shao2024deepseekmath}.

\section{Proposed \names Framework}

\subsection{Problem Overview}\label{sec:prob_overview}

Fig. ~\ref{fig:ADS} overviews our problem. We consider human subjective tasks (Fig.~\ref{fig:ADS}a) where annotations are derived from human perceptions, interpretations, and opinions. Labels are collected from multiple annotators, and we treat ambiguity that arises between them as information, not noise.

Assume the labels collected from $N$ human annotators for sample $q$ are denoted as:
\begin{equation}
\small
    \{l_{(q,1)}, l_{(q,2)}, \dots, l_{(q,N)}\},
\end{equation}

\noindent where each $l_{(q,i)} \in \{1, \dots, C\}$ corresponds to one of $C$ candidate classes. 

Instead of collapsing annotations into a single majority label, we derive an \textit{Empirical Label Distribution} over annotator judgment, which serves as the supervision signal:

\begin{equation}\label{eq:p_qc}
\small
p_{q,c}
=
\frac{n_{q,c}}{N},
\quad
c = 1, \dots, C,
\quad
\sum_{c=1}^{C} p_{q,c}=1,
\end{equation}

\noindent where $n_{q,c}$ denotes the number of annotators assigning class $c$ for sample $q$, and $\mathbf p_q \in \mathbb R^C$ represents the corresponding annotator distribution.

%Given an input sample $x_q$, the language model generates a sampled output $o_{(q,i)}$ associated with a predicted probability distribution over candidate classes: \paul{this part is unclear does the LM generate tokens or classify labels?}

Given an input sample $x_q$, the LLM generates a sampled textual response $o_{(q,i)}$ containing predicted probabilities over the candidate label space:
\begin{equation}\label{eq:p_hat}
\small
\hat{\mathbf p}_{(q,i)}
=
[
\hat p_{(q,i),1},
\hat p_{(q,i),2},
\dots,
\hat p_{(q,i),C}
].
\end{equation}

\begin{equation}
\small
\hat p_{(q,i),c}
=
P(\hat y = c \mid x_q, o_{(q,i)}).
\end{equation}

We formulate the learning objective within a reward-based optimization framework (Fig.~\ref{fig:ADS}b). Instead of optimizing toward a single collapsed label, our goal is to \textit{encourage generated predictions that better align with the underlying distribution of human interpretations}. Based on this formulation, we develop \names %r an \textbf{\textit{Ambiguity-Enhanced Distributional Supervision (\names)}} paradigm 
that enables LLMs to learn directly from annotator distributions under ambiguous human annotations. We adopt GRPO~\citep{shao2024deepseekmath} as the optimization backbone due to its flexibility in reward design and its effectiveness in LLM alignment. This formulation allows %r ambiguity-aware distributional supervision
smart handling of ambiguous labels
to be naturally incorporated in reward-based learning.

\subsection{Verbalized Distributions from LLM}\label{sec:verbalize}

To obtain distributional outputs from LLMs, we adopt a verbalized distribution prediction formulation (Fig.~\ref{fig:ADS}.b-a), where the model directly generates probability estimates over candidate classes in textual form. Such verbalized distributions have been shown to provide an effective mechanism for eliciting uncertainty-aware predictions from LLMs~\cite{radharapu2025arbiters}.

%Given an input sample $x_q$, the language model generates a sampled textual response $o_{(q,i)}$ that verbalizes the predicted distribution in natural language form:
The sampled textual response $o_{(q,i)}$ from  the input sample $x_q$ is represented as a verbalized distribution over candidate classes:

%\vspace{-1em}
\begin{equation}
\small
o_{(q,i)}=
\{
(c_1, \hat p_{(q,i),1}),
%(c_2, \hat p_{(q,i),2}),
\dots,
(c_C, \hat p_{(q,i),C})
\},
\end{equation}

\noindent where $\hat p_{(q,i),c}$ denotes the predicted probability assigned to candidate class $c$ for rollout $o_{(q,i)}$. The resulting probability estimates are then parsed into the predicted label distribution $\hat{\mathbf p}_{(q,i)}$ defined in Eq.~(\ref{eq:p_hat}).

%Unlike conventional single-label prediction, verbalized distributions explicitly preserve uncertainty and relative confidence across candidate classes. This formulation enables ambiguity-aware reward optimization by directly aligning model-predicted distributions with annotator disagreement distributions under subjective supervision.

\subsection{Group Relative Policy Optimization}\label{sec:grpo}
%\paul{this should go earlier into a background section not part of the method}
%\jw{@Keane, could you work on 3.1.2? This part just include standard GRPO and its equations etc, and + one sentence about why we think GRPO is a good fit. }
%To enable language models to learn directly from annotator distributions under ambiguous supervision, 
%We formulate the learning objective within a reward-based optimization framework. Instead of optimizing toward a single collapsed label, our goal is to encourage generated predictions that better align with the underlying distribution of human interpretations. We adopt Group Relative Policy Optimization (GRPO)~\citep{shao2024deepseekmath} as the optimization backbone due to its flexibility in reward design and its effectiveness in LLM alignment. This formulation allows ambiguity-aware distributional supervision to be naturally incorporated through reward-based learning.

%We initialize a policy $\pi_\theta(a\mid s)$, as a Large Langauge Model \keane{put your specific backbone model LLM}. To optimize this policy, we consider Group Relative Policy Optimization (GRPO)~\citep{shao2024deepseekmath}, a recent on-policy RL method that has shown strong performance for reasoning-based LLM training. For task $m$ and sample $q$, GRPO samples a rollout group $G_{(m,q)}$ of responses $\{o_{(m,q,i)}\}$, where $i \in G_{(m,q)}$ indexes individual rollouts (i.e., a sampled response) with rewards $r_{(m,q,i)}$, computing the group-normalized advantage:
We consider an LLM parameterized by $\theta$ as the policy model $\pi_\theta(a \mid s)$, where $s$ denotes the input prompt and $a$ denotes the generated output. Following GRPO~\citep{shao2024deepseekmath}, for each sample $q$, a rollout group $G_{(q)}$ consisting of multiple sampled outputs $\{o_{(q,i)}\}$ is generated, where $i \in G_{(q)}$ indexes individual rollouts with corresponding rewards $r_{(q,i)}$. GRPO then computes the group-normalized advantage:
 %Therefore the reward formulations can then be incorporated into GRPO optimization to encourage varying forms of alignment with human disagreement. %\paul{$\mathcal{R}$ is not defined}
%We consider a large language model parameterized by $\theta$ as the policy model $\pi_\theta(a \mid s)$, where $s$ denotes the input prompt and $a$ denotes the generated response. Following GRPO, for task $m$ and sample $q$, a rollout group $G_{(m,q)}$ consisting of multiple sampled responses $\{o_{(m,q,i)}\}$ is generated, where $i \in G_{(m,q)}$ indexes individual rollouts with corresponding rewards $r_{(m,q,i)}$. GRPO then computes the group-normalized advantage:

%GRPO then computes the group-normalized advantage:
\begin{equation}
\small
\label{eq:grpo_adv}
\hat A_{(q,i)}
=
\frac{r_{(q,i)}-\hat\mu_{G_{(q)}}}{\hat\sigma_{G_{(q)}}+\varepsilon}
\end{equation}
where $\hat\mu_{G_{(q)}}$ and $\hat\sigma_{G_{(q)}}$ are the mean
and standard deviation of $\{r_{(q,i)}\}_{i=1}^{|G_{(q)}|}$. With a PPO clipped surrogate \( \tilde A_{(q,i):k}(\theta) \) constructed from \( \hat A_{(q,i)} \), GRPO then optimizes $\pi_\theta(a\mid s)$ using a PPO-style trust-region objective\footnote{Full formulation of GRPO is provided in App.~\ref{app:grpo_full_formulation}.}:
\begin{equation}
\small
\begin{aligned}
J_{\mathrm{GRPO}}(\theta)
&=
\mathbb{E}_{(q)\sim \mathcal D}
\mathbb{E}_{\{o_{(q,i)}\}\sim \pi_{\theta_{\mathrm{old}}}}\!\Bigg[
\frac{1}{|G_{(q)}|}
\sum_{i\in G_{(q)}}
\\
&\hspace{-60pt}\phantom{=\mathbb{E}\Bigg[}
\frac{1}{n_{o_{(q,i)}}}
\sum_{k=1}^{n_{o_{(q,i)}}}
\tilde A_{(q,i):k}(\theta)
\Bigg]
\;-\;
\beta\,
\mathbb{E}\!\left[
D_{\mathrm{KL}}\!\left(\pi_\theta\;\|\;\pi_{\mathrm{ref}}\right)
\right]
\end{aligned}
\end{equation}

%\jw{but also need to balance what's to be included here, and what should be included in the appendix, keane I think you're good at this since you've written many AI conferences.}

%\jw{Maybe can start with ...}We adopt Group Relative Policy Optimization (GRPO) as the underlying optimization framework for training language models with reward-based supervision.

%+paragraph 1: what is GRPO (introduce grpo conceptually)
%+paragraph 2: Standard GRPO objective.

%\jw{and end with} In this work, we retain the standard GRPO optimization procedure and focus on reward design under ambiguous supervision.

\subsection{\namesr~Reward}

%Traditional supervised learning commonly collapses multiple human annotations into a single majority label, thereby discarding potentially informative disagreement structures across annotators. 
Our hypothesis is that ambiguous samples often contain disagreement structures that provide informative signals regarding nuance, interpretational diversity, uncertainty, and conflicting human perceptions. We present a novel way to use this hypothesis to improve LLM alignment: \namess, and describe its %r:\namesm%\textbf{\textit{Ambiguity-Enhanced Distributional Supervision (\names)}} 
reward-based optimization algorithm within GRPO.

\names consists of two key components (Fig.~\ref{fig:ADS}b-b): First, it aligns LLM predictions with annotator distributions, preserving the structure of human judgments during optimization rather than collapsing supervision into a dominant label. Second, it %r incorporates ambiguity-aware enhanced modulation that 
dynamically reweights rollout contributions according to the degree of annotator disagreement, enabling highly ambiguous samples to exert greater influence during policy optimization.

%\paragraph{\emph{\namesr} reward.}

%r Within the proposed \emph{\names} paradigm and 
Following the GRPO formulation in Section~\ref{sec:grpo}, for each sampled rollout output $o_{(q,i)}$ generated from the LM, we define the ambiguity-enhanced reward:
\begin{equation}
\small
r_{(q,i)}^{\mathrm{\namesr}}
=
\tilde H(\mathbf p_q)
\left[
1
-
D_{\mathrm{JS}}
\left(
\hat{\mathbf p}_{(q,i)},
\mathbf p_q
\right)
\right],
\end{equation}

\noindent where $\hat{\mathbf p}_{(q,i)}$ denotes the predicted label distribution generated from rollout $o_{(q,i)}$ following Section~\ref{sec:verbalize}, and $\mathbf p_q$ denotes the corresponding annotator distribution for sample $q$ defined in Section~\ref{sec:prob_overview}.

%\paul{are you sure this has not been done before? defining distribution similarity as reward?} \jw{to my knowledge, i didn't see any works like this, certainly no in ER; and based on the ChaoNLI benchmark papers and the distributional alignment papers, they were not using anything related to RL.} 

The component,
$
1
-
D_{\mathrm{JS}}
(
\hat{\mathbf p}_{(q,i)},
\mathbf p_q
)
$,
measures agreement between the predicted distribution and the annotator distribution using Jensen-Shannon (JS) Distance. Since $D_{\mathrm{JS}}(\cdot,\cdot)\in[0,1]$, the resulting reward is also bounded within $[0,1]$.

%\paul{this part seems new. but it is not emphasize much.. should explain more} %The component,$\tilde H(\mathbf p_q)$, introduces ambiguity-aware reward modulation based on the normalized entropy of the annotator distribution:

Importantly, not all disagreement structures are equally informative during learning. To explicitly prioritize highly ambiguous samples containing richer uncertainty and interpretational diversity, we introduce ambiguity-enhance reward modulation based on the normalized entropy of the annotator distribution:
\begin{equation}
\small
\tilde H(\mathbf p_q)
=
\frac{
-
\sum_{c=1}^{C}
p_{q,c}\log p_{q,c}
}{
\log C
},
\end{equation}

\noindent where $\tilde H(\mathbf p_q)\in[0,1]$ quantifies the degree of annotator disagreement in sample $q$ while $p_qc$ refers to the empirical label distribution defined in (\ref{eq:p_qc}).

\paragraph{Distributional alignment under annotator ambiguity.}
%\paul{disagreement/ambiguity/subjectivity pls choose one and stick with it}
%The proposed reward is incorporated into the GRPO framework through the rollout reward term in Eq.~(\ref{eq:grpo_adv}). Since GRPO computes policy updates through group-normalized rollout advantages, the structure of the reward distribution directly influences rollout contribution dynamics. Unlike discrete majority-label rewards that induce sparse and highly concentrated reward signals, \names produces continuous rewards proportional to distributional agreement with human annotations. A benefit of this is that sampled outputs exhibiting partial alignment with annotator distributions can still contribute positively to policy learning. This reshapes the relative reward geometry within each rollout group, yielding smoother advantage distributions and more informative policy updates under ambiguous supervision. Compared with conventional majority-label optimization, \names preserves the full structure of human judgments, enabling the model to utilize information from diverse and potentially conflicting human labels rather than only the dominant view.
The proposed reward is incorporated into the GRPO framework through the rollout reward term in Eq.~(\ref{eq:grpo_adv}). Since GRPO computes policy updates using group-normalized rollout advantages, the reward distribution directly influences rollout contribution dynamics. Unlike discrete majority-label rewards that produce sparse and highly concentrated signals, \names generates continuous rewards proportional to distributional agreement with human annotations. As a result, sampled outputs exhibiting partial alignment with annotator distributions can still contribute positively to policy learning. This reshapes the reward geometry within each rollout group, yielding smoother advantage distributions and more informative policy updates under ambiguous supervision. Compared with conventional majority-label optimization, \names preserves the full structure of human judgments, enabling the model to learn from diverse and potentially conflicting labels rather than only the dominant view.

\paragraph{Dynamic ambiguity-modulated policy optimization.}

While distributional alignment preserves the full structure of human judgments, \names further introduces %r ambiguity-enhanced 
reward modulation that dynamically reweights rollout contributions utilizing annotator ambiguity. Since GRPO computes policy updates through rollout advantage magnitudes, \namesr~ reward scaling directly modulates the relative contribution strength of sampled outputs. Samples with higher annotator ambiguity receive proportionally amplified rollout rewards and therefore exert greater influence on the resulting policy updates. Compared with uniform distributional supervision, the ambiguity-conditioned optimization by \namesr~ %r introduces ambiguity-conditioned optimization emphasis that prioritizes learning from highly ambiguous samples during training, 
potentially enables the policy model to better learn variability %rand diverse human interpretations 
while maintaining alignment with the underlying annotator distributions.

\section{Experimental Setup}

\paragraph{Datasets.} We evaluate the proposed \names\ framework on ambiguity-sensitive human-centered tasks exhibiting different forms of annotator disagreement, including semantic ambiguity in NLI and affective ambiguity in ER. For NLI, we conduct experiments on ChaosNLI \cite{nie-etal-2020-learn}, which contains 100 annotations per sample to capture diverse semantic interpretations and disagreement, as well as its underlying ChaosNLI-M \cite{mnli-2018} and ChaosNLI-S \cite{snli-2015} subsets. For ER, we evaluate on MSP-Podcast \cite{msp-podcast}, one of the largest naturalistic speech emotion corpora, and GoEmotions \cite{2020-goemotions}, which contains diverse emotion categories and intentionally ambiguous samples. Both emotion datasets include 5--12 annotations per sample, reflecting subjective emotional perception and interpretation. For MSP-Podcast, we include both speech and text modalities to evaluate \names\ under multimodal ambiguity.

\paragraph{Model.} All experiments are conducted using Qwen2.5-Omni-7B~\cite{Qwen2.5-Omni} as the LLM. We adopt a unified GRPO-based optimization framework across all experiments following the GRPOTrainer provided by TRL \cite{trl2020}.

\paragraph{Baselines.} We compare the proposed \names against a range of baselines and ablation settings. These include: (1) a Zero-Shot (ZS) inference model without task-specific training; (2) Majority-Label Supervision (MLS) (refer to App.~\ref{app:mls}), which encourages the model to assign high probability mass to the dominant annotation and serves as a reward-based analogue of conventional majority-label training; (3) recent state-of-the-art methods reported for each dataset, as justified in App.~\ref{app:stoa}; and (4) an ablation variant without ambiguity enhancement, denoted as \names (w/o) Ambi-En, which removes reward modulation by setting $\tilde H(\mathbf p_q)=1$.

\paragraph{Evaluation.} To evaluate distributional alignment between model predictions and annotator distributions, we report Jensen-Shannon Distance (JSD) ($\downarrow$) and Bhattacharyya Coefficient (BC) ($\uparrow$). Both metrics lie in [0,1]. We additionally report conventional classification metrics, including Accuracy ($\uparrow$), macro F1-score ($\uparrow$) and Weighted F1 (W-F1) ($\uparrow$) to assess whether distributional supervision maintains competitive performance under standard evaluation protocols. We note that the latter metrics do not account for ambiguity\footnote{Full details of experimental settings are in App.~\ref{app:experiment}}. %We additionally analyze model performance under varying ambiguity levels to better understand learning behavior under annotator disagreement.
%Details of the experimental setups can be found in App. \ref{app:experiment}.
\section{Results and Discussion}
\subsection{Overall Performance Comparison}

\begin{table*}[tb!]
\centering
\footnotesize
\setlength{\tabcolsep}{3pt}

\caption{Performance comparisons of the proposed \textbf{\names} framework with baselines across NLI and ER datasets. Relative percentage changes are computed with respect to Zero-shot inference (ZS). Best and second-best results are highlighted in \textbf{bold} and \underline{underline}. Relative performance improvement (+) and degradation (-) compared with the ZS are reported in brackets.} %\textcolor{blue}{Relative performance improvement (+) and degradation (-) compared with the ZS are reported in brackets.}}
\label{tab:baseline_results}

\resizebox{\textwidth}{!}{
\begin{tabular}{llccccc}
\toprule
Dataset & Method & JSD ($\downarrow$) & BC ($\uparrow$) & Acc ($\uparrow$) & F1 ($\uparrow$) & W-F1 ($\uparrow$) \\
\midrule

\multirow{4}{*}{ChaosNLI}
& Zero-shot & 0.375 & 0.850 & 0.603 & 0.473 & 0.547 \\
& MLS & 0.477 (-27.2\%) & 0.751 (-11.6\%) & 0.699 (+15.9\%) & 0.650 (+37.4\%) & 0.684 (+25.0\%) \\
& \names~(w/o Ambi-En) & \underline{0.192 (+48.8\%)} & \underline{0.964 (+13.4\%)} & \underline{0.736 (+22.1\%)} & \underline{0.686 (+45.0\%)} & \underline{0.721 (+31.8\%)} \\
& \textbf{\names} & \textbf{0.181 (+51.7\%)} & \textbf{0.966 (+13.6\%)} & \textbf{0.768 (+27.4\%)} & \textbf{0.758 (+60.3\%)} & \textbf{0.767 (+40.2\%)} \\
\midrule

\multirow{4}{*}{ChaosNLI-M}
& Zero-shot & 0.376 & 0.845 & 0.649 & 0.511 & 0.607 \\
& MLS & 0.510 (-35.6\%) & 0.731 (-13.5\%) & 0.695 (+7.1\%) & \underline{0.637 (+24.7\%)} & \underline{0.683 (+12.5\%)} \\
& \names~(w/o Ambi-En) & \underline{0.188 (+50.0\%)} & \underline{0.968 (+14.6\%)} & \underline{0.701 (+8.0\%)} & 0.588 (+15.1\%) & 0.671 (+10.5\%) \\
& \textbf{\names} & \textbf{0.173 (+54.0\%)} & \textbf{0.972 (+15.0\%)} & \textbf{0.760 (+17.1\%)} & \textbf{0.737 (+44.2\%)} & \textbf{0.758 (+24.9\%)} \\
\midrule

\multirow{4}{*}{ChaosNLI-S}
& Zero-shot & 0.375 & 0.855 & 0.557 & 0.436 & 0.490 \\
& MLS & 0.445 (-18.7\%) & 0.770 (-9.9\%) & 0.703 (+26.2\%) & 0.654 (+50.0\%) & 0.686 (+40.0\%) \\
& \names~(w/o Ambi-En) & \underline{0.195 (+48.0\%)}  & \underline{0.960 (+12.3\%)} & \underline{0.769 (+38.1\%)} & \underline{0.746 (+71.1\%)} & \underline{0.763 (+55.7\%)} \\
& \textbf{\names} & \textbf{0.191 (+49.1\%)} & \textbf{0.961 (+12.4\%)} & \textbf{0.775 (+39.1\%)} & \textbf{0.767 (+75.9\%)} & \textbf{0.775 (+58.2\%)} \\
\midrule

\multirow{4}{*}{\shortstack{MSP-Podcast\\(Speech + Text)}}
& Zero-shot & 0.640 & 0.508 & 0.421 & 0.266 & 0.388 \\
& MLS & 0.580(+10.3\%) & 0.585(+7.7\%) & \underline{0.488(+13.7\%)} & 0.233(-3.3\%) & 0.415(+2.7\%) \\
& \names~(w/o Ambi-En) & \underline{0.550 (+14.1\%)} & \underline{0.658 (+29.5\%)} & 0.482 (+14.5\%) & \underline{0.276 (+3.8\%)} & \underline{0.433 (+11.6\%)} \\
& \textbf{\names} & \textbf{0.544(+15.0\%)} & \textbf{0.694(+36.7\%)} & \textbf{0.496(+17.8\%)} & \textbf{0.301(+13.2\%)} & \textbf{0.455 (+17.3\%)} \\
\midrule

\multirow{4}{*}{\shortstack{GoEmotions\\(Text)}}
& Zero-shot & 0.681 & 0.480 & 0.361 & 0.345 & 0.377 \\
& MLS & 0.542(+20.4\%) & 0.638(+32.9\%) & 0.595(+64.8\%) & \textbf{0.591(+71.3\%)} & \underline{0.595(+57.8\%)} \\
& \names~(w/o Ambi-En) & \textbf{0.449 (+34.1\%)} & \underline{0.750 (+56.3\%)} & \textbf{0.611 (+69.3\%)} & {0.544 (+57.7\%)} & \textbf{0.603 (+59.9\%)} \\
& \textbf{\names} & 0.465 (+31.7\%) & \textbf{0.756 (+57.5\%)} & \underline{0.600(+66.2\%)} & \underline{0.589(+70.7\%)} & \underline{0.595(+57.8\%)} \\
\bottomrule
\end{tabular}
}
\end{table*}

\begin{table}[tb!]
\centering
\scriptsize
\setlength{\tabcolsep}{2pt}

\caption{Performance comparison between the new \textbf{\names} framework and existing ambiguity-aware approaches across NLI and ER datasets. Best results are highlighted in \textbf{bold}.}
\label{tab:main_results}

\begin{tabular}{llccccc}
\toprule
Dataset & Method & JSD$\downarrow$ & BC$\uparrow$ & Acc$\uparrow$ & F1$\uparrow$ & WF1$\uparrow$ \\
\midrule

\multirow{3}{*}{ChaosNLI}
& \makecell[l]{LLM-Explain\\\cite{chen-etal-2025-rose}} & 0.207 & -- & -- & -- & 0.645 \\
& \makecell[l]{LLM-MJD\\\cite{chen-etal-2024-seeing}} & 0.208 & -- & -- & -- & 0.621 \\
& \textbf{\names} & \textbf{0.181} & \textbf{0.966} & \textbf{0.768} & \textbf{0.758} & \textbf{0.767} \\
\midrule

\multirow{5}{*}{ChaosNLI-M}
& \makecell[l]{Chaos-Benchmark\\\cite{nie-etal-2020-learn}} & 0.306 & -- & 0.635 & -- & -- \\
& \makecell[l]{Dist. NLI\\\cite{distnli}} & 0.197 & -- & 0.637 & -- & -- \\
& \makecell[l]{AmbiNLI\\\cite{ambinli}} & 0.223 & -- & 0.584 & -- & -- \\
& \makecell[l]{Flan-T5\\\cite{lee-etal-2023}} & 0.260 & -- & 0.726 & -- & -- \\
& \textbf{\names} & \textbf{0.173} & \textbf{0.972} & \textbf{0.760} & \textbf{0.737} & \textbf{0.758} \\
\midrule

\multirow{5}{*}{ChaosNLI-S}
& \makecell[l]{Chaos-Benchmark\\\cite{nie-etal-2020-learn}} & 0.220 & -- & 0.787 & -- & -- \\
& \makecell[l]{Dist. NLI\\\cite{distnli}} & \textbf{0.181} & -- & \textbf{0.794} & -- & -- \\
& \makecell[l]{AmbiNLI\\\cite{ambinli}} & 0.189 & -- & 0.755 & -- & -- \\
& \makecell[l]{Flan-T5\\\cite{lee-etal-2023}} & 0.231 & -- & 0.738 & -- & -- \\
& \textbf{\names} & 0.191 & \textbf{0.961} & 0.775 & \textbf{0.767} & \textbf{0.775} \\
\midrule

\multirow{2}{*}{MSP-Podcast}
& \makecell[l]{TTS-Benchmark\\\cite{jia2026decoding}} & \textbf{0.285} & 0.621 & 0.423 & 0.253 & -- \\
& \textbf{\names} & 0.544 & \textbf{0.649} & \textbf{0.496} &\textbf{0.301} & \textbf{0.455}\\
\midrule

\multirow{3}{*}{GoEmotions}
& \makecell[l]{AER-LLM (ZS)\\\cite{aer-llm}} & {0.49} & 0.54 & 0.371 & -- & 0.357 \\
& \makecell[l]{AER-LLM (FS)\\\cite{aer-llm}} & \textbf{0.44} & 0.70 & 0.505 & -- & 0.511 \\
& \textbf{\names} &{0.47} & \textbf{0.76} & \textbf{0.60} & \textbf{0.59} & \textbf{0.60}  \\
\bottomrule
\end{tabular}%\vspace{-2em}

\end{table}

Tables~\ref{tab:baseline_results} and \ref{tab:main_results} show the performance across ambiguity-sensitive tasks under both distributional and conventional single-label evaluation metrics. %The proposed \names framework consistently outperforms conventional baselines. \names achieves competitive or superior performance compared with previously reported ambiguity-aware approaches across both NLI and emotion recognition tasks, demonstrating the effectiveness of ambiguity-aware distributional supervision under human disagreement.

\paragraph{Comparison with baselines.} As shown in Table~\ref{tab:baseline_results}, \names consistently improves both distributional alignment and conventional classification performance compared with ZS and MLS across all datasets and evaluation metrics. On ChaosNLI, \names reduces JSD from 0.477 to 0.181 (62.1\% $\downarrow$) while improving BC from 0.751 to 0.966 (28.6\% $\uparrow$) compared with MLS. \names also substantially improves conventional classification performance, increasing ACC from 0.699 to 0.768 (9.9\% $\uparrow$), F1 from 0.650 to 0.758 (16.6\% $\uparrow$), and W-F1 from 0.684 to 0.767 (12.1\% $\uparrow$). Strong consistent trends are also observed in its subset ChaosNLI-M and ChaosNLI-S and observed across ER tasks. %where \names achieves substantial improvements in both distributional alignment and single-label evaluation performance compared with ZS. 
On MSP-Podcast, \names improves BC from 0.580 to 0.544 (6.2\%$\uparrow$) and F1 from 0.233 to 0.301 (29.2\%$\uparrow$) compared to MLS. Although the performance differences on GoEmotion between \names and MLS remain relatively moderate on conventional classification metrics, the overall results demonstrate that \names remains highly effective across ambiguous ER settings. %On GoEmotions, \names further demonstrates strong gains across both distributional and classification metrics, substantially improving JSD, BC, ACC, F1, and W-F1 compared with baseline approaches. 
Overall, these findings suggest that preserving annotator disagreement during optimization not only improves alignment with human judgment distributions, but can also benefit dominant-label prediction performance under ambiguous supervision.

Notably, we observe for NLI datasets, although the conventional MLS generally improves conventional classification performance over ZS, it simultaneously leads to substantially worse distributional alignment. For example, on ChaosNLI, MLS increases JSD from 0.374 to 0.477 (27.2\% $\uparrow$) while reducing BC from 0.851 to 0.751 (11.6\% $\downarrow$). These findings suggest that MLS encourages the model to collapse ambiguity into dominant labels, thereby discarding variability and disagreement structures embedded within annotator distributions.

\paragraph{Ablation studies.} We further compare \names against its ablation variant, \names (w/o) Ambi-En, which removes ambiguity-enhanced reward modulation and optimizes only the distributional alignment objective. While \names(w/o) already demonstrates strong improvements over ZS and MLS, the full \names framework consistently achieves further gains across both distributional and classification metrics. On ChaosNLI, \names further reduces JSD from 0.192 to 0.181 (5.4\% $\downarrow$) while improving BC from 0.964 to 0.966 (0.2\% $\uparrow$) compared to \names w/o Ambi-En. \names also improves ACC from 0.736 to 0.768 (4.3\% $\uparrow$), F1 from 0.686 to 0.758 (10.5\% $\uparrow$), and W-F1 from 0.721 to 0.767 (6.4\% $\uparrow$). 

On GoEmotions, although the performance differences remain relatively moderate, the full \names framework still demonstrates competitive and consistently stronger performance across several evaluation metrics compared with \names (w/o) Ambi-En, including improvements on BC (0.8\%$\uparrow$) and F1 (8.3\%$\uparrow$). These findings suggest that ambiguity-enhanced reward modulation provides additional optimization benefits beyond distributional alignment alone, while the primary gains are already largely achieved through ambiguity-aware distributional supervision.

%Similar improvements are consistently observed across other datasets and evaluation metrics. 

\paragraph{Comparison with state of the art.} %\paul{how is this different from comparison with baselines 2 paragraphs ago?} \jw{This paragraph is comparing with recent works (table 2). Previous baselines are ZS and MLS and abalation (table 1)} 
%We further compare \names with previously reported ambiguity-aware approaches and benchmark results across multiple datasets. While direct comparison should be interpreted with caution due to differences in model architectures and experimental settings, \names achieves competitive or superior performance on both distributional and conventional metrics. For example, on ChaosNLI, \names achieves lower JSD and substantially higher W-F1 compared with recent ambiguity-aware methods including LLM-Explain~\cite{chen-etal-2025-rose} and LLM-MJD~\cite{chen-etal-2024-seeing}. Similar trends are also observed on MSP-Podcast and GoEmotions.%, where \names demonstrates strong robustness across diverse contextual and semantic reasoning settings. 
%These findings further support the effectiveness and generalizability of ambiguity-aware distributional optimization across diverse ambiguity-sensitive tasks.
\begin{figure*}[!tb]
\centering
\includegraphics[width=\linewidth,height = 3.3cm]{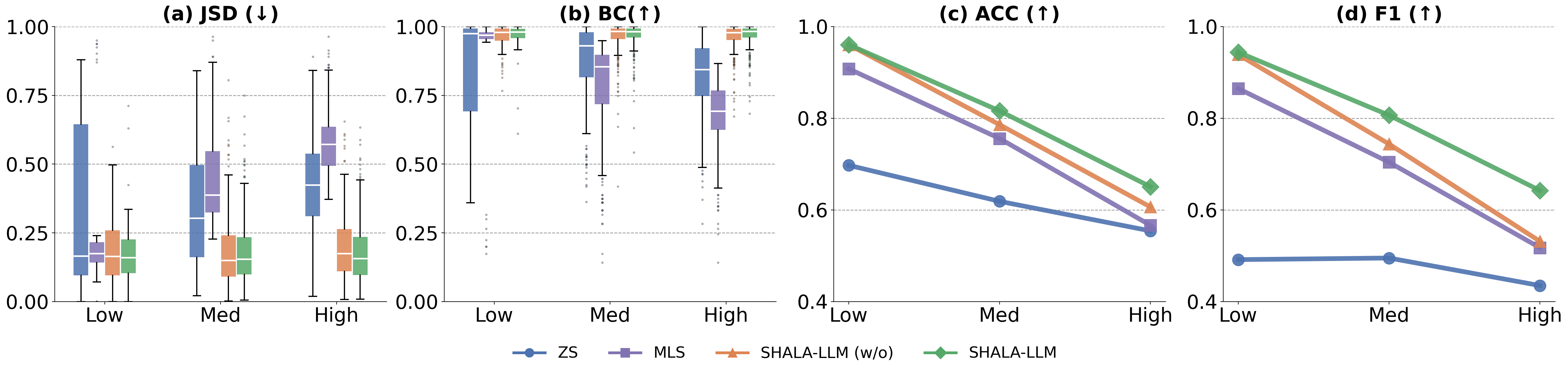}
\caption{Performance comparison across different ambiguity levels on the ChaosNLI dataset.}
\label{fig:Chaos_ambi_level}
\end{figure*}

We further compare \names with previously reported ambiguity-aware approaches. While direct comparisons should be interpreted cautiously due to differences in model architectures and experimental settings, \names achieves competitive or superior performance on both distributional and conventional metrics. For example, on ChaosNLI, \names achieves lower JSD and substantially higher W-F1 than recent ambiguity-aware methods, including LLM-Explain~\cite{chen-etal-2025-rose} and LLM-MJD~\cite{chen-etal-2024-seeing}. Similar trends are observed on MSP-Podcast and GoEmotions. These findings further support the effectiveness and generalizability of \namess~ across diverse ambiguity-sensitive tasks.

%Collectively, these findings suggest that highly ambiguous samples provide particularly informative supervision signals during optimization. By dynamically prioritizing samples exhibiting substantial annotator disagreement, \names encourages the model to better capture uncertainty and diverse human interpretations embedded within ambiguous annotations. Overall, the results demonstrate that incorporating ambiguity-aware optimization into reinforcement learning based alignment leads to stronger alignment with human judgment distributions under subjective supervision.

Collectively, these findings suggest that highly ambiguous samples provide particularly informative supervision signals during optimization. By dynamically prioritizing samples with substantial annotator disagreement, \names\ encourages the model to better capture uncertainty and diverse human interpretations embedded in ambiguous annotations. Overall, the results demonstrate that ambiguity-aware optimization in reinforcement learning–based alignment leads to stronger agreement with human judgment distributions under subjective supervision.

\subsection{Analysis at Different Ambiguity Levels}\label{sec:ambi_level}

To gain deeper insight into the robustness and effectiveness of ambiguity-aware optimization, we further analyze model performance across varying levels of annotator disagreement. Specifically, we partition samples according to the entropy of annotator distributions and evaluate different supervision strategies under low-, medium-, and high-ambiguity settings. Details of the data partition procedure can be found in App.\ref{app:dataset}.

\paragraph{Overall performance at all ambiguity levels.} Fig.~\ref{fig:Chaos_ambi_level} presents the results on the ChaosNLI dataset. Overall, the proposed \names framework consistently achieves superior performance compared with ZS and MLS baselines across all ambiguity levels. Specifically, for distributional evaluation (Fig.~\ref{fig:Chaos_ambi_level}a-b), both \names and its ablation variant \names(w/o) exhibit consistently better median performance together with reduced variance across ambiguity levels compared with baseline methods. Similar observations are also seen for conventional classification metrics (Fig.~\ref{fig:Chaos_ambi_level}c-d), where \names (green lines) consistently achieves the strongest overall performance across all ambiguity levels.

\paragraph{Robustness as ambiguity level increases.}

%Importantly, horizontally across the boxplots, the performance of \names and \names(w/o) remains comparatively stable as ambiguity levels increase, with no statistically significant degradation observed across ambiguity levels ($p > 0.05$). In contrast, the baseline methods exhibit substantially larger performance degradation under higher ambiguity conditions. On the line plots, while all methods naturally exhibit lower performance under highly ambiguous settings, the degradation remains relatively smaller under \names. For example, the F1 of \names drops 31.97\% from low- to high-ambiguity, whereas MLS exhibits a substantially larger drop of 40.17\%. This suggests that preserving annotator disagreement during optimization improves robustness under increasingly subjective supervision conditions.

Importantly, across the boxplots, the performance of \names and \names(w/o) remains relatively stable as ambiguity increases, with no statistically significant degradation observed across ambiguity levels ($p > 0.05$). In contrast, the baseline methods exhibit substantially larger degradation under higher ambiguity conditions. Similarly, although all methods show lower performance in highly ambiguous settings, the degradation remains smaller for \names. For example, the F1 of \names decreases by 32.0\% from low- to high-ambiguity, whereas MLS shows a larger drop of 40.2\%. These findings suggest that preserving annotator disagreement during optimization improves robustness under increasingly subjective supervision conditions.

% \paragraph{Robustness as ambiguity level increases.} Importantly, horizontally across the boxplots, it is observed that the performance of AmbED and AmbED (w/o) remains comparatively stable as ambiguity levels increase, evidenced by that with no significant changes across three levels ($p<0.1$). Whereas the baseline methods exhibit substantially larger performance degradation under higher ambiguity conditions. On the line plots, while all methods naturally exhibit lower performance under highly ambiguous settings, the degradation remains relatively smaller under AmbED. This suggests that preserving annotator disagreement during optimization improves robustness under increasingly subjective supervision conditions.

Interestingly, MLS (purple boxplots) demonstrates competitive and occasionally superior performance under low-ambiguity settings where dominant consensus labels are clearer. However, its performance drops substantially under medium- and high-ambiguity conditions, for example, its BC decreases by 28.56\% from low (0.970) to high-ambiguity (0.693) with even worse performance at high-ambiguity level compared to ZS (0.844). This suggests that conventional majority-label optimization struggles to generalize once supervision becomes increasingly subjective and disagreement structures become more prominent.

%f1 ambed 31.97 \%  (0.9441 to 0.6423)
%f1 mls  40.17\%     (0.8648 to 0.5174)

%Finally, comparing AmbED against its ablation variant AmbED (w/o), we observe that AmbED consistently provides additional robustness improvements under higher ambiguity levels, evidenced by its overall higher performance, and smaller performance drops as ambiguity level increases. This indicates that dynamically prioritizing highly ambiguous samples during optimization further strengthens the model’s ability to learn from rich disagreement structures embedded within human annotations.
\begin{figure*}[!tb]
\centering
\includegraphics[width=\linewidth,height = 3.3cm]{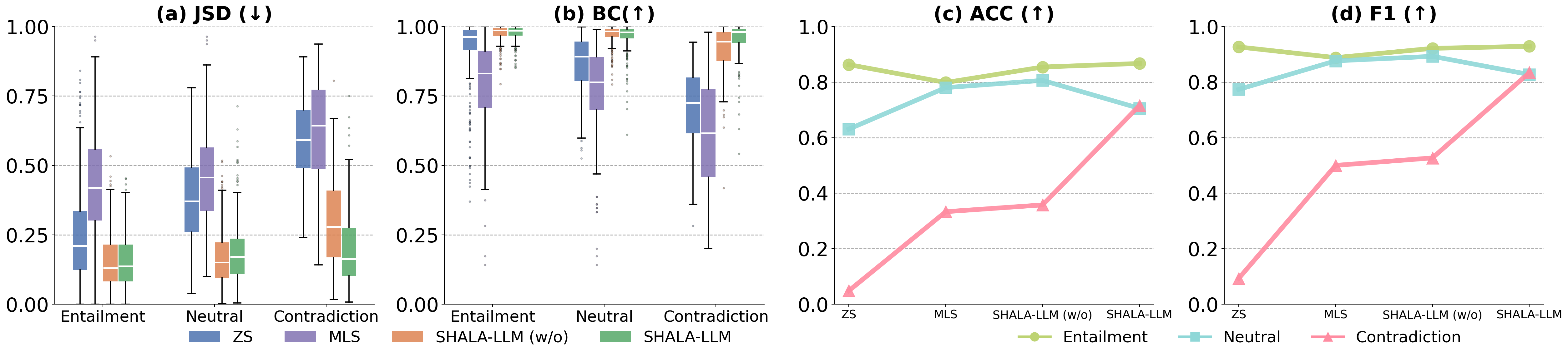}
\caption{Category-level performance analysis on the ChaosNLI dataset across different classes.}\vspace{-1em} %\paul{the figure is hard to read. axes and text too small. too many subplots} \jw{I still think we can include it since we have space?} \jw{@Ashley, update figure model names to SHALALLM}}
\label{fig:contradiction}
\end{figure*}

Finally, comparing \names against its ablation variant \names(w/o), we observe that \names consistently achieves stronger robustness under higher ambiguity levels, reflected by both higher overall performance and smaller performance degradation as ambiguity increases. These findings indicate that dynamically prioritizing highly ambiguous samples during optimization further strengthens the model’s ability to learn from rich disagreement structures embedded within human annotations.

\subsection{Interpreting Model Behavior under Ambiguous Supervision}\label{sec:model_reasoning}

% \begin{figure*}[!tb]
% \centering
% \includegraphics[width=\linewidth]{figures/nli_majority_class_row.png}
% \caption{Category-level performance analysis on the ChaosNLI dataset across different classes. \paul{this graph is not the most important} }
% \label{fig:contradiction}
% \end{figure*}

%We further investigate how ambiguity-aware optimization influences model behavior under ambiguous supervision. Specifically, we analyze both category-level performance patterns and representative reasoning examples to better understand how \names captures disagreement structures and multiple plausible interpretations under ambiguous conditions.
We further investigate how ambiguity-aware optimization influences model behavior under ambiguous supervision. We analyze category-level performance patterns and representative reasoning examples to better understand how \names captures disagreement structures and multiple plausible interpretations in ambiguous settings.
\paragraph{Quantitative analysis across different semantic categories.}

Fig.~\ref{fig:contradiction} presents a category-level analysis on the ChaosNLI dataset across the \textit{Entailment}, \textit{Neutral}, and \textit{Contradiction} classes. Overall, both \names and its ablation variant \names (w/o) consistently improve performance across all semantic categories compared with ZS and MLS baselines. The largest improvements are observed for the \textit{Contradiction} class, which also exhibits the highest ambiguity and annotator disagreement.

While all methods achieve relatively strong performance on \textit{Entailment} and \textit{Neutral}, baseline approaches struggle substantially on \textit{Contradiction} samples. In contrast, \names achieves dramatic improvements across all evaluation metrics for contradiction cases (Fig.~\ref{fig:contradiction}c-d), improving F1 from 0.09 (ZS) and 0.50 (MLS) to 0.83, while ACC increases from 0.05 (ZS) and 0.33 (MLS) to 0.72. Similar trends are also observed on distributional metrics with substantially lower JSD and higher BC metrics.

%These findings suggest that ambiguity-aware optimization provides particularly strong benefits for semantic categories involving richer semantic uncertainty and conflicting interpretations. By preserving disagreement structures during optimization, \names better captures multiple plausible semantic relationships under ambiguous supervision.

These findings suggest that ambiguity-aware optimization is particularly beneficial for semantic categories with greater uncertainty and conflicting interpretations. By preserving disagreement structures during optimization, \names better captures multiple plausible semantic relationships under ambiguous supervision.

%\paragraph{Interpreting Reasoning Behavior under High Ambiguity.} We further examine the reasoning behaviours of the model under \names training. %We observe that AmbED often produces distributions that more closely align with human disagreement patterns, while simultaneously generating reasoning that better reflects multiple plausible interpretations and underlying uncertainty. Additional low- and medium-ambiguity examples are provided in the Appendix.
%Table~\ref{tab:chaos_case} presents a representative highly ambiguous example from the ChaosNLI dataset. While the ZS baseline produces an overconfident prediction with complete certainty, \names generates both probability estimates and reasoning patterns that more closely reflect the underlying annotator disagreement (refer to the bold text in the table). %In particular, AmbED explicitly acknowledges that ``there are multiple valid interpretations'' within the premise-hypothesis relationship, whereas the ZS baseline instead asserts that there are ``no multiple valid interpretations.'' 
%Rather than collapsing the sample into a single definitive interpretation, \names reasons probabilistically over multiple plausible semantic relationships, resulting in a predicted distribution that better aligns with human judgment distributions. These findings suggest that ambiguity-enhanced optimization influences not only final predictions, but also the underlying reasoning behavior of the model under subjective supervision.

\paragraph{Interpreting reasoning behavior under high ambiguity.} We further examine the reasoning behavior of the model under \names training. Table~\ref{tab:chaos_case} presents a representative highly ambiguous example from the ChaosNLI dataset. While the ZS baseline gives an overconfident prediction, \names generates probability estimates and reasoning patterns that more closely reflect the underlying annotator disagreement (refer to the bold text in the table). Rather than collapsing the sample into a single interpretation, \names reasons over multiple plausible semantic relationships, producing a predicted distribution that better aligns with human judgment distributions. These findings suggest that ambiguity-aware optimization influences not only final predictions, but also the model’s underlying reasoning behavior.
% under subjective supervision.

\begin{table}[tb!]
\scriptsize
\centering
\caption{\textbf{Premise}: "In short, we all got tired of clever analyses of what might happen; and throughout economics there was a shift in focus away from theorizing, toward data collection and careful statistical analysis"; \textbf{Hypothesis}: "We all got tired of data collection and clever analyses of what might happen";Labels: Entailment (E), Neutral (N), or Contradiction (C).} %\jw{@Ashely, I shorten the model resoning in this table, can you put the full version to appendix?}}
\label{tab:chaos_case}
\setlength{\tabcolsep}{2pt}
\renewcommand{\arraystretch}{1.1}

\begin{tabularx}{\columnwidth}{|p{0.11\columnwidth}|X|p{0.05\columnwidth}|p{0.05\columnwidth}|p{0.05\columnwidth}|}
\hline
\textbf{} & \textbf{Model Reasoning} & \textbf{E} & \textbf{N} & \textbf{C} \\
\hline

Ground Truth & & .45 & .17 & .38 \\
\hline

%MLS & Model didn't output any reasoning. & 1.00 & .00 & .00 \\
%\hline

Zero-shot & 
The premise and hypothesis are in direct contradiction. %The premise suggests a shift towards data collection, while the hypothesis suggests a shift away from data collection. 
$\dots$ There are \textbf{no multiple valid interpretations} here. $\dots$ %The premise and hypothesis are clearly in conflict. $\dots$

%Since the premise and hypothesis are in direct contradiction, the probability of entailment is \textbf{0}. The probability of neutrality is also \textbf{0} because the relationship is not neutral. The probability of contradiction is \textbf{1} because the hypothesis directly contradicts the premise.

& .00 & .00 & 1.00 \\

\hline

\names &

\textbf{There are multiple valid interpretations}: 

1. The premise and hypothesis could be seen as contradictory because $\dots$ %the premise indicates a shift away from clever analyses, while the hypothesis suggests tiredness with both data collection and clever analyses. 
2. Alternatively, the premise could be interpreted as a shift $\dots$ %in focus, which might imply that the tiredness with clever analyses was a result of the shift towards data collection and statistical analysis. 

Given these interpretations, the probability of entailment is relatively \textbf{low}, as... The probability of neutrality is \textbf{moderate}, as... The probability of contradiction is \textbf{high}, as...

& .20 & .50 & .30 \\

\hline
\end{tabularx}\vspace{-1em}
\end{table}

\section{Conclusion}
In this work, we introduced \namesm~(\namess), a reinforcement learning based alignment framework that enables large language models to learn directly from annotator distributions under ambiguous supervision. By preserving disagreement structures and dynamically prioritizing highly ambiguous samples during optimization, \names consistently improves both distributional alignment and conventional classification performance across natural language inference and emotion recognition tasks, while demonstrating stronger robustness under highly ambiguous conditions. Overall, our findings highlight the value of treating human disagreement and ambiguity as informative supervision signals rather than annotation noise, suggesting a promising direction toward more robust and human-centered large language models under subjective real-world settings.

\section*{Limitations}
One limitation of this work is that \names is primarily evaluated on tasks with relatively structured label distributions. %r and moderate-scale ambiguity settings. 
While our approach demonstrates improved alignment with distributions of human judgments, the current experiments do not yet capture more complex forms of ambiguity that arise in open-ended generative tasks, long-form reasoning, or real-world multi-turn interactions. In addition, our reward formulation assumes that annotator disagreement can be adequately represented through empirical label distributions, which may not fully reflect underlying factors such as annotator expertise, demographic variation, or contextual uncertainty. Moreover, our experiments on Qwen2.5-Omni-7B disabled stochastic sampling to improve output consistency and parsing stability for verbalized distribution generation. Future work should further investigate the uncertainty of model-generated distributions and how sampling variability may influence ambiguity-aware optimization under subjective supervision. Finally, although we focus on GRPO-based reasoning alignment, the generalization of the proposed ambiguity-aware reward design to other reinforcement learning and preference optimization frameworks remains %r underexplored and will require further investigation in
future work.

\section*{Ethical Considerations}

This work focuses on ambiguity-aware alignment for subjective human-centered tasks, including natural language inference and emotion recognition, where disagreement between annotators often reflects diverse interpretations and perceptions rather than annotation noise. By preserving annotator disagreement distributions during optimization, the proposed framework aims to better capture uncertainty and interpretational diversity instead of enforcing a single dominant label. Nevertheless, the underlying datasets may still contain societal, cultural, or demographic biases inherited from human annotations, which could influence model behavior and potentially amplify biased disagreement patterns. Our framework partially mitigates this issue by explicitly modeling disagreement distributions rather than collapsing annotations into majority labels, thereby preserving diverse interpretations under subjective supervision. Additionally, outputs from ambiguity-aware models should not be interpreted as definitive representations of human emotions, intentions, or beliefs, particularly in high-stakes decision-making settings.

Potential risks of this work include the possibility that societal, cultural, or demographic biases embedded within human annotations may also be preserved or amplified through ambiguity-aware optimization. Additionally, outputs from ambiguity-aware models could be misinterpreted as definitive representations of human emotions, intentions, or beliefs, particularly in high-stakes decision-making settings. Our framework partially mitigates these risks by explicitly modeling disagreement distributions rather than collapsing annotations into majority labels, thereby preserving diverse interpretations and uncertainty under subjective supervision.

%This work focuses on ambiguity-aware alignment for subjective human-centered tasks, including natural language inference and emotion recognition, where disagreement between annotators often reflects diverse interpretations and perceptions rather than annotation noise. By preserving annotator disagreement distributions during optimization, the proposed framework aims to better capture uncertainty and interpretational diversity instead of enforcing a single dominant label. Nevertheless, the underlying datasets may still contain societal, cultural, or demographic biases inherited from human annotations, which could influence model behavior and potentially amplify biased disagreement patterns. Additionally, outputs from ambiguity-aware models should not be interpreted as definitive representations of human emotions, intentions, or beliefs, particularly in high-stakes decision-making settings.

\section*{Acknowledgments}

\bibliography{custom}

\appendix

\clearpage

%\section{Limitations}\label{sec:limitations}
%\subsection{Research Limitations}
%\subsection{Potential Risks}
%\subsection{Ethical Considerations}
\section{Full GRPO Formulation}\label{app:grpo_full_formulation}
For completeness, we provide the full formulation of GRPO, including the surrogate objective and importance-sampling formulation. Accordingly, for task $m$ and sample $q$, GRPO samples a rollout group $G_{(m,q)}$ of responses $\{o_{(m,q,i)}\}$, where $i \in G_{(m,q)}$ indexes individual rollouts (i.e., a sampled response) with rewards $r_{(m,q,i)}$, computing the group-normalized advantage:
\begin{equation}
\label{eq:grpo_adv2}
\hat A_{(m,q,i)}
=
\frac{r_{(m,q,i)}-\hat\mu_{G_{(m,q)}}}{\hat\sigma_{G_{(m,q)}}+\varepsilon},
\end{equation}
where $\hat\mu_{G_{(m,q)}}$ and $\hat\sigma_{G_{(m,q)}}$ are the empirical mean
and standard deviation of $\{r_{(m,q,i)}\}_{i=1}^{|G_{(m,q)}|}$. GRPO then optimizes $\pi_\theta(a\mid s)$ by performing a PPO-style trust-region update. At token position $k$ of response $o_{(m,q,i)}$,
$\varphi_{(m,q,i):k}(\theta)$ denotes the importance sampling ratio between
$\pi_\theta$ and the old policy $\pi_{\theta_{\mathrm{old}}}$,
$\tilde A_{(m,q,i):k}(\theta)$ denotes the PPO-clipped surrogate using
$\hat A_{(m,q,i)}$, and $J_{\mathrm{GRPO}}(\theta)$ averages this surrogate over
tokens and rollout samples with an optional KL penalty to a reference policy
$\pi_{\mathrm{ref}}$ (with weight $\beta$). We summarize these with a compact
objective:
\begin{equation}
\small
\label{eq:grpo_compact2}
\begin{aligned}
\varphi_{(m,q,i):k}(\theta)
&=
\frac{
\pi_\theta\!\big(o_{(m,q,i):k}\mid q,\, o_{(m,q,i):<k}\big)
}{
\pi_{\theta_{\mathrm{old}}}\!\big(o_{(m,q,i):k}\mid q,\, o_{(m,q,i):<k}\big)
}
\\
\tilde A_{(m,q,i):k}(\theta)
&=
\min\!\Big(
\varphi_{(m,q,i):k}(\theta)\,\hat A_{(m,q,i)},
\\
& \hspace{-50pt} \phantom{=\min\!\Big(}
\operatorname{clip}\!\big(\varphi_{(m,q,i):k}(\theta),\,1-\epsilon,\,1+\epsilon\big)\,\hat A_{(m,q,i)}
\Big)
\\
J_{\mathrm{GRPO}}(\theta)
&=
\mathbb{E}_{q\sim\mathcal D_m}
\mathbb{E}_{\{o_{(m,q,i)}\}\sim \pi_{\theta_{\mathrm{old}}}}\!\Bigg[
\frac{1}{|G_{(m,q)}|}
\sum_{i=1}^{|G_{(m,q)}|}
\\
&\hspace{-75pt}\phantom{=\mathbb{E}\Bigg[}
\frac{1}{n_{o_{(m,q,i)}}}
\sum_{k=1}^{n_{o_{(m,q,i)}}}
\tilde A_{(m,q,i):k}(\theta)
\Bigg]
\;-\;
\beta\,
\mathbb{E}\!\left[
D_{\mathrm{KL}}\!\left(\pi_\theta\;\|\;\pi_{\mathrm{ref}}\right)
\right].
\end{aligned}
\end{equation}
\section{Dataset Description}\label{app:dataset}

% \jw{I just moved the datasets from main paper to here, you probly need to write more details about dataset statistics here}
% \jw{model reasoning explananation output of specific examples, for each level, 10 from low , medium, high level etc}

\subsection{ChaosNLI Dataset}
ChaosNLI \cite{nie-etal-2020-learn} is a text-based dataset designed to capture human disagreement in natural language inference tasks. It consists of premise–hypothesis pairs drawn from established NLI benchmarks and reannotated with 100 independent human raters per example. Raters are asked to determine whether the hypothesis is entailed by, contradicted by, or neutral with respect to the premise. In this work, we restrict our analysis to examples originating from the Stanford Natural Language Inference Corpus (SNLI) \cite{snli-2015} and Multi-Genre Natural Language Inference Corpus (MNLI) \cite{mnli-2018}. 

ChaosNLI-S \cite{snli-2015} is a large-scale natural language inference dataset constructed from image captions in the Flickr30k corpus. Human annotators are presented with a text caption describing an image and asked to write three corresponding hypotheses: one that is definitely true given the caption (entailment), one that may be true but is uncertain (neutral), and one that is definitely false (contradiction). This process produces sentence pairs with relatively simple linguistic structure and forces the data to be balanced among these classes.

ChaosNLI-M \cite{mnli-2018} is modeled on the SNLI corpus but extends the task to a broader range of spoken and written text. It collects premise sentences from ten distinct genres including fiction, government reports, travel guides, telephone speech, and spoken conversations. Annotators are then asked to label hypothesis–premise pairs according to the same entailment, neutral, and contradiction categories, resulting in a dataset with more diverse language and more challenging inference phenomena.

% \jw{can we split SNLI and MNLI as another two separate datasets?}
\subsection{MSP Podcast}
The MSP Podcast (v1.12) \cite{msp-podcast} dataset is a large-scale naturalistic speech corpus constructed from podcast audio. Each short audio segment is annotated by at least five human raters for perceived emotion labels. To ensure consistency across samples, we follow prior work in restricting the label space to a standard set of eight emotion categories: anger, contempt, disgust, fear, happiness, neutral, sadness, and surprise \cite{aer-llm, wu2026amber}. Examples containing annotations outside this label set are discarded. 

\subsection{GoEmotions}
GoEmotions \citep{2020-goemotions} is a large-scale text emotion dataset consisting of English Reddit comments annotated with fine-grained emotion labels. Each example is initially annotated by 3--5 human raters, where annotators are allowed to assign multiple emotion labels to a single example. If no two annotators agree on any emotion label, two additional raters are assigned to the example. Annotators may also choose to assign no label if they consider the example too difficult to classify. Rather than using the aggregated binary labels released with the original dataset, we utilize the raw annotator-level responses to construct human annotation distributions.

We apply several preprocessing steps before constructing these distributions. First, we discard examples where fewer than three annotators provide valid emotion labels to ensure a reliable estimate of an underlying distribution. Second, the original dataset contains 27 fine-grained emotion categories and a neutral label, making distribution prediction substantially more difficult and sparse. Following the Ekman taxonomy proposed in \citep{2020-goemotions}, we map the original labels into six broader emotion categories as shown in Table~\ref{tab:ekman_mapping}. For each example, annotator labels are mapped to the corresponding Ekman categories, aggregated across raters, and normalized to produce a probability distribution over emotions.

Finally, analysis of the majority label of resulting annotation distributions revealed substantial class imbalance. We observed that a large proportion of examples were concentrated in the joy and neutral classes. To reduce bias toward high-frequency classes and improve evaluation across emotions, we apply stratified sampling to select approximately equal numbers of examples for each dominant emotion category.

\begin{table}[h]
\centering
\small
\caption{Mapping of GoEmotions 27 fine-grained emotion labels to Ekman's six basic emotion categories \citep{2020-goemotions}.}
\begin{tabular}{p{1.5cm}p{5.5cm}}
\toprule
\textbf{Ekman Category} & \textbf{GoEmotions Fine-Grained Labels} \\
\midrule
Joy      & Admiration, Amusement, Approval, Caring, Desire, Excitement, Gratitude, Joy, Love,\\
         &  Optimism, Pride, Relief \\
\midrule
Sadness  & Disappointment, Embarrassment, Grief, Remorse, Sadness \\
\midrule
Anger    & Anger, Annoyance, Disapproval \\
\midrule
Surprise & Confusion, Curiosity, Realization, Surprise \\
\midrule
Fear     & Fear, Nervousness \\
\midrule
Disgust  & Disgust \\
\midrule
Neutral  & Neutral \\
\bottomrule
\end{tabular}
\label{tab:ekman_mapping}
\end{table}

\section{Data Processing}
\subsection{Dataset statistics}
\subsubsection{ChaosNLI}
The ChaosNLI dataset consists of 3,113 examples, including 1,514 examples sourced from SNLI and 1,599 examples sourced from MNLI. Prior to training, the dataset is randomly shuffled and partitioned into training, validation, and test sets using a 65/15/20 split. Table~\ref{tab:chaosNLI_stats} summarizes the distribution of the majority label across the dataset, where the majority label for an example is defined as the label receiving the largest proportion of annotations.

\begin{table}[h]
\centering
\caption{ChaosNLI dataset statistics by majority label.}
\label{tab:chaosNLI_stats}
\begin{tabular}{llrr}
\toprule
\textbf{Dataset} & \textbf{Majority Label} & \textbf{Count} \\
\midrule
\multirow{4}{*}{ChaosNLI}
  & Entailment     & 1{,}168  \\
  & Neutral        & 1{,}397 \\
  & Contradiction  &   548  \\
\midrule
\multirow{4}{*}{ChaosNLI-S}
  & Entailment     &  424 \\
  & Neutral        &  811  \\
  & Contradiction  &  279 \\
\midrule
\multirow{4}{*}{ChaosNLI-M}
  & Entailment     &  744  \\
  & Neutral        &  586 \\
  & Contradiction  &  269  \\
\bottomrule
\end{tabular}
\end{table}

\subsubsection{MSP Podcast}
After preprocessing and filtering, the resulting dataset contains 12,955 examples. Following prior work \cite{wu2026amber}, we adopt an even five-fold partitioning strategy rather than the standard speaker-based splits to achieve a more balanced distribution of ambiguous examples across data partitions. We divide the dataset into five equal-sized folds, using three folds for training, one for validation, and one for testing, resulting in an approximate 60/20/20 split. Table~\ref{tab:msp_stats} summarizes the distribution of majority emotion labels across the dataset. 

\begin{table}[h]
\centering
\caption{MSP Podcast dataset statistics by majority label.}
\label{tab:msp_stats}
\begin{tabular}{lr}
\toprule
\textbf{Majority Label} & \textbf{Count} \\
\midrule
  Angry     & 1,053  \\
   Contempt        & 1053 \\
  Disgust  &   806  \\
  Fear     & 603  \\
 Happy        & 3519 \\
   Neutral  &   4609  \\
   Sad     & 432  \\
   Surprise        & 880 \\
\bottomrule
\end{tabular}
\end{table}
\subsubsection{GoEmotions}
After preprocessing and filtering, we collect 6567 examples in the GoEmotions dataset. Examples are randomly shuffled and partitioned into training, validation, and test sets using a 70/15/15 split. Table~\ref{tab:goemo_stats} summarizes the distribution of the majority emotion label across the dataset.

\begin{table}[h]
\centering
\caption{GoEmotions dataset statistics by majority label.}
\label{tab:goemo_stats}
\begin{tabular}{lr}
\toprule
\textbf{Majority Label} & \textbf{Count} \\
\midrule
  Anger     & 1,000  \\
   Disgust        & 1000 \\
  Fear  &   640  \\
  Joy     & 927  \\
 Sadness        & 1000 \\
   Surprise  &   1000  \\
   Neutral     & 1000  \\
\bottomrule
\end{tabular}
\end{table}

\subsection{Data partition across different ambiguity levels} \label{app:amb_level}

In addition to overall dataset statistics, we analyze how examples are distributed across different levels of ambiguity. 

\subsubsection{ChaosNLI}
For ChaosNLI, ambiguity is quantified using the entropy of the ground-truth annotation distribution. Lower entropy corresponds to stronger agreement among annotators, while higher entropy indicates greater disagreement. We partition examples into three ambiguity levels: low, medium, and high, using normalized entropy thresholds of $[0,0.33)$, $[0.33,0.66)$, and $[0.66,1]$, respectively. Table~\ref{tab:chaosNLI_amb_stats} summarizes the resulting distribution of examples across these ambiguity levels.

\begin{table}[h]
\centering
\caption{ChaosNLI dataset statistics by ambiguity level.}
\label{tab:chaosNLI_amb_stats}
\begin{tabular}{llrr}
\toprule
\textbf{Dataset} & \textbf{Ambiguity Level} & \textbf{Count} \\
\midrule
\multirow{4}{*}{ChaosNLI}
  & Low     & 377  \\
  & Medium        & 1493 \\
  & High  &   1243  \\
\midrule
\multirow{4}{*}{ChaosNLI-S}
  & Low     &  327 \\
  & Medium        &  865  \\
  & High  &  322 \\
\midrule
\multirow{4}{*}{ChaosNLI-M}
  & Low     &  50  \\
  & Medium        &  628 \\
  & High  &  921  \\
\bottomrule
\end{tabular}
\end{table}

An interesting observation is that ChaosNLI-M contains a larger proportion of examples concentrated in the high-ambiguity level compared to ChaosNLI-S. This difference likely reflects characteristics of the underlying source datasets. ChaosNLI-S is constructed from image captions and therefore tends to contain shorter, more concrete descriptions with relatively straightforward relationship structures. In contrast, ChaosNLI-M spans a broader range of genres and linguistic styles, introducing more complex language and inference patterns that naturally lead to greater annotator disagreement.

\subsubsection{MSP-Podcast}\label{app:msp_ambi}
For MSP-Podcast, we characterize ambiguity using the number of active labels present in the ground-truth distribution rather than entropy directly. We define an active label as an emotion category receiving non-zero probability mass in the annotation distribution. Intuitively, the lowest ambiguity occurs when all annotators agree on a single emotion label. Examples with two active labels correspond to cases where annotators only identify two distinct emotion categories, while progressively larger numbers of active labels indicate greater disagreement among raters. We partition examples into categories of one, two, three, and four-or-more active labels. Table~\ref{tab:msp_amb_stats} presents the distribution of examples across these groups along with the average normalized entropy of the ground-truth distribution.

\begin{table}[h]
\centering
\caption{MSP Podcast dataset statistics by ambiguity level.}
\label{tab:msp_amb_stats}
\begin{tabular}{llr}
\toprule
\textbf{Active Labels} & \textbf{Count} & \textbf{Average Entropy} \\
\midrule
1 label &2963 & 0.0000\\
2 labels& 2410& 0.2537\\
3 labels& 3152& 0.4661\\
4+ labels& 4430&  0.6604 \\
\bottomrule
\end{tabular}
\end{table}

\subsubsection{GoEmotions}\label{app:goemo_ambi}
Similar to MSP-Podcast, ambiguity in GoEmotions is characterized using the number of active labels within the ground-truth distribution. Examples with a single active label represent strong annotator agreement, whereas examples with multiple active labels indicate increasing diversity in annotation responses. We group examples into categories containing one, two, three, and four-or-more active labels. Table~\ref{tab:goemo_amb_stats} summarizes the distribution of examples across these ambiguity levels and the average entropy of the corresponding ground-truth distributions.

\begin{table}[h]
\centering
\caption{GoEmotions dataset statistics by ambiguity level.}
\label{tab:goemo_amb_stats}
\begin{tabular}{llr}
\toprule
\textbf{Active Labels} & \textbf{Count} & \textbf{Average Entropy} \\
\midrule
1 label & 1434 & 0.0000\\
2 labels& 1828& 0.3140\\
3 labels& 1714& 0.5138\\
4+ labels& 1591&  0.6875 \\
\bottomrule
\end{tabular}
\end{table}

\section{Experimental Setup}\label{app:experiment}

\subsection{Artifact Usage and Licensing.}
All datasets and pretrained models used in this work are publicly available for research purposes and were used in accordance with their intended usage conditions and licenses.

\subsection{Model and GRPO Configuration}
%\jw{C1 Model Size And Budget*
%Did you report the number of parameters in the models used, the total computational budget (e.g., GPU hours), and computing infrastructure used?}
%\jw{C4 Parameters For Packages*
%If you used existing packages (e.g., for preprocessing, for normalization, or for evaluation, such as NLTK, SpaCy, ROUGE, etc.), did you report the implementation, model, and parameter settings used?}

%\keane{btw ashley, could you double check the batch sizes, because 1 batch size over 2 steps may be quite small. Also, please check if you can specify the mini batch, micro batch configs. Did you also include the KL divergence parameters? See if you want to also spell out the GRPOConfigs here, for easier reading.}

% \ashley{double checked, i mostly followed configurations from other paper \cite{rouditchenko2025omnir1} that also did grpo with qwen omni2.5, kl divergence beta=0 by default in grpotrainer }

%\jw{that's fine, you can mention we follow the same configuration as that paper due to xxx +add one short sentence of why.}
%\jw{one thing, Roz mentioned: we should also mention we set do\_sample = false in the Qwen configuration, which give us deterministic decoding and generating tokens with the highest probability at each step and it is reproducible, follow the settings in + refer to that rouditchenko2025omnir1 paper}

We conduct all experiments using the Qwen2.5-Omni-7B model~\cite{Qwen2.5-Omni} and implement training with the GRPOTrainer from the TRL framework~\cite{trl2020}. We follow the GRPO configuration settings of \cite{rouditchenko2025omnir1}, which also performs GRPO fine-tuning on Qwen2.5-Omni, particularly for improving the model's audio interpretation capabilities. We utilize the AdamW optimizer with an initial learning rate of $1\times10^{-6}$. Following the GRPO setup, we set the number of rollouts to 4, temperature to 1.2, and maximum completion length to 128 tokens. We additionally set $\beta = 0$, removing the KL divergence regularization term from the original GRPO formulation \cite{shao2024deepseekmath}. This choice is motivated by recent studies \cite{liu2025r1zeroliketraining,hu2025zeroopensourceapproach} showing that the KL divergence term is not necessary effective GRPO training. All experiments were conducted using the Qwen2.5-Omni-7B model backbone, which contains approximately 7 billion parameters.

For generation, we set do\_sample to False, resulting in deterministic decoding where the model selects the highest-probability token at each generation step. This improves reproducibility and ensures that evaluation results are not affected by sampling variation. Training is performed on a single compute node with two NVIDIA H200 GPUs and 400GB of memory. The batch size per GPU is set to 1 with gradient accumulation over two steps, resulting in an effective batch size of 4 prompts per optimization step. We employ DeepSpeed ZeRO Stage 3 optimization for efficient distributed training and memory management. Training progress and evaluation metrics are logged using WandB. Unless otherwise specified, all remaining hyperparameters follow the default settings defined in GRPOConfig within TRL~\cite{trl2020}. Results are reported as single-run experiments due to the computational cost of GRPO-based LLM fine-tuning.

%\subsection{GRPO Configuration}
\subsection{Prompts}
Below, we provide the prompt templates used for training across each dataset. Table~\ref{tab:nli_prompt} presents the prompt template used for ChaosNLI, Table~\ref{tab:msp_prompt} shows the template for MSP Podcast, and Table~\ref{tab:goemotions_prompt} shows the template used for GoEmotions. While the task-specific context and label spaces differ across datasets, all prompts follow a common structure consisting of background information, the target utterance to evaluate, task instructions, and output formatting constraints.

\begin{table}[h]
\centering
\scriptsize
\setlength{\tabcolsep}{3pt}
\renewcommand{\arraystretch}{1.1}
\caption{Prompt template used for ChaosNLI with example target utterance.}
\begin{tabularx}{\linewidth}{p{0.18\linewidth} X}
\toprule
\textbf{Section} & \textbf{Prompt Content} \\
\midrule

\textbf{Background} &
 \\
\midrule

\textbf{Target Utterance} &
Premise: A child in a red jacket, waist deep in a pit on the beach

Hypothesis: A child is building a
sandcastle on the beach \\
\midrule
\textbf{Task} &
Given a premise and a hypothesis, predict the probability of the relationship between them from the following options: entailment, neutral, contradiction.
\begin{enumerate}
    \item entailment: the hypothesis logically follows from the premise
    \item neutral: neither entailment nor contradiction can be determined
    \item contradiction: the hypothesis conflicts with the premise
\end{enumerate}

You MUST produce a calibrated probability distribution. \\
\midrule

\textbf{Output Constraints} &
Before outputting, check if the format of your output is in accordance with the requirements I provided. 
\begin{enumerate}
    \item 1. Generate the label probabilities in EXACTLY this JSON structure: \{\{"entailment": float, "neutral": float, "contradiction": float\}\}.
    
    \item The sum of all probabilities must be exactly 1.0.
    
    \item Do not include any explanations or text besides the dictionary.
\end{enumerate}
\\
\bottomrule
\end{tabularx}
\label{tab:nli_prompt}
\end{table}

\begin{table}[h]
\centering
\scriptsize
\setlength{\tabcolsep}{3pt}
\renewcommand{\arraystretch}{1.1}
\caption{Prompt template used for MSP Podcast with example target utterance.}
\begin{tabularx}{\linewidth}{p{0.18\linewidth} X}
\toprule
\textbf{Section} & \textbf{Prompt Content} \\
\midrule

\textbf{Background} &
Two speakers are having a conversation. \\
\midrule

\textbf{Target Utterance} &
that's right. spilling my load of liberty all over your faces. it's the golden stallion of the tech- \\
\midrule
\textbf{Task} &
Predict the probability of the emotion in the target utterance from the following options: angry, contempt, disgust, fear, happy, neutral, sad, surprise. 

You MUST produce a calibrated probability distribution. 
Identify any emotional cues expressed by the speaker, even subtle ones. For each of the following emotions from angry, contempt, disgust, fear, happy, neutral, sad, surprise, evaluate whether the emotion is present. If only one emotion present, assign 1.0 to that emotion. If multiple emotions are present, estimate the relative strength and assign a probability to each emotion. The probability should reflect how much time, intensity, and presence each emotion has in the conversation.  \\
\midrule

\textbf{Output Constraints} &
Before outputting, check if the format of your output is in accordance with the requirements I provided. 
\begin{enumerate}
    \item Generate the emotion probabilities in EXACTLY this JSON structure: \{\{"Angry": float, "Contempt": float, "Disgust": float, "Fear": float, "Happy": float, "Neutral": float, "Sad": float, "Surprise": float\}\}.
    
    \item The sum of all probabilities must be exactly 1.0.
    
    \item Do not include any explanations or text besides the dictionary.
\end{enumerate}
\\
\bottomrule
\end{tabularx}
\label{tab:msp_prompt}
\end{table}

\begin{table}[h]
\centering
\scriptsize
\setlength{\tabcolsep}{3pt}
\renewcommand{\arraystretch}{1.1}
\caption{Prompt template used for GoEmotions with example target utterance.}
\begin{tabularx}{\linewidth}{p{0.18\linewidth} X}
\toprule
\textbf{Section} & \textbf{Prompt Content} \\
\midrule

\textbf{Background} &
This is a text comment extracted from Reddit. \\
\midrule

\textbf{Target Utterance} &
Dear [NAME] man! (Irony intended) You didn\'t say "proof" you said "evidence"! \\
\midrule
\textbf{Task} &
Assign a calibrated probability distribution over the following emotion categories: anger, disgust, fear, joy, sadness, surprise, neutral.

These categories follow the Ekman taxonomy:
\begin{itemize}
\item anger: annoyance, disapproval, hostility
\item disgust: contempt, revulsion
\item fear: nervousness, anxiety, dread
\item joy: happiness, admiration, gratitude, excitement, love, optimism, pride, relief, amusement
\item sadness: disappointment, grief, remorse, embarrassment
\item surprise: curiosity, confusion, realization
\item neutral: no clear emotion expressed
\end{itemize}

Predict the probability of the emotion in the target utterance from the following options: anger, disgust, fear, joy, sadness, surprise, neutral.

You MUST produce a calibrated probability distribution.

For each of the following emotions from anger, disgust, fear, joy, sadness, surprise, neutral, evaluate whether the emotion is present. If only one emotion present, assign 1.0 to that emotion. If multiple emotions are present, estimate the relative strength and assign a probability to each emotion. The probability should reflect how much intensity and presence each emotion has in the comment. \\
\midrule

\textbf{Output Constraints} &
Before outputting, check if the format of your output is in accordance with the requirements I provided. 
\begin{enumerate}
    \item Generate the emotion probabilities in EXACTLY this JSON structure: \{\{"anger": float, "disgust": float, "fear": float, "joy": float, "sadness": float, "surprise": float, "neutral": float\}\}.
    
    \item The sum of all probabilities must be exactly 1.0.
    
    \item Do not include any explanations or text besides the dictionary.
\end{enumerate}
\\
\bottomrule
\end{tabularx}
\label{tab:goemotions_prompt}
\end{table}

\subsection{Majority-label supervision (MLS) Baseline}\label{app:mls}
As a reference baseline, we first consider conventional Majority-label supervision (MLS), where the annotator distribution is reduced to its majority label:

\begin{equation}
y_q^* = \arg\max_c p_{q,c},
\end{equation}

\noindent where $p_{q,c}$ denotes the annotator distribution defined in Eq.~\ref{eq:p_qc}.

%The rollout reward is then defined as the probability assigned to the majority label:
Given the predicted distribution $\hat{\mathbf p}_{(q,i)}$ defined in Eq.~\ref{eq:p_hat}, the rollout reward is defined as the probability assigned to the majority label:
\begin{equation}
r_{(q,i)}^{\mathrm{maj}}
=
\hat p_{(q,i),y_q^*}.
\end{equation}

This formulation encourages the model to assign high probability mass to the dominant annotation and serves as a reward-based analogue of conventional MLS.

\subsection{Justification of Included Baselines}\label{app:stoa}

We include a diverse range of recent studies and strong baselines for comparison across both ambiguity-aware learning and large language model alignment settings. On ChaosNLI (including ChaosNLI-M and ChaosNLI-S), we include the original benchmark framework proposed together with the dataset collection process, denoted as Chaos-Benchmar~\cite{nie-etal-2020-learn}. We further include recent studies that explicitly model annotator disagreement or distributional supervision under the same benchmark settings, including distribution-aware LLM alignment and ambiguity-aware learning approaches~\cite{chen-etal-2025-rose, chen-etal-2024-seeing, lee-etal-2023}. 

Additionally, we include prior ambiguity-aware fine-tuning methods such as~\cite{distnli, ambinli}, which are based on BERT-style architectures rather than LLM backbones. Although these methods do not perform reinforcement learning based alignment, they are included because they optimize directly on annotator distributions and report evaluation results using the same distributional metrics adopted in our work, enabling meaningful comparison under ambiguity-aware supervision.

For emotion recognition datasets, there remain relatively limited prior studies explicitly addressing ambiguity-aware distributional supervision. On MSP-Podcast, we include the recent benchmark study on decoding ambiguous emotions using test-time scaling, denoted as TTS-Benchmark~\cite{jia2026decoding}. On GoEmotions, we include recent LLM-based ambiguity-aware emotion recognition approaches~\cite{aer-llm} that report distributional emotion outputs under both zero-shot (ZS) and few-shot (FS) settings, enabling direct comparison under ambiguity-aware evaluation protocols.

\section{Evaluation Metrics}\label{app:metrics}
\textbf{Jensen-Shannon Distance} Jensen-Shannon distance (JS) is a metric that measures the similarity between two probability distributions. It is a symmetric and smoothed version of the Kullback–Leibler divergence (KL) with finite values bounded between 0 and 1 when using base 2 logarithms. A smaller JS distance would indicate more similarity between two probability distributions, with zero indicating two identical distributions. Given probability distributions $P$ and $Q$, let $M=\frac{P+Q}{2}$ and $D_{\mathrm{KL}}(P \| Q)$ be the KL divergence given by 
\begin{equation*}
    D_{\mathrm{KL}}(P \| Q) = \sum_{x \in \mathcal{X}} P(x) \log \frac{P(x)}{Q(x)}.
\end{equation*}
The JS distance is defined as
\begin{equation*}
    D_{\mathrm{JS}}(P \| Q) = \sqrt{\frac{1}{2} D_{\mathrm{KL}}(P \| M) + \frac{1}{2} D_{\mathrm{KL}}(Q \| M)}.
\end{equation*}
By comparing comparing both distributions to a central mixture distribution, the JS distance metric is more stable than KL divergence. In our analysis, we set $Q$ to be the ground truth probability distribution from the annotations and $P$ to be the model's predicted probability distribution. \\ \\
\textbf{Bhattacharyya Coefficient} The Bhattacharyya Coefficient (BC) measures the amount of overlap or statistical similarity between two probability distributions. It is bounded between 0 and 1, where a value of 1 indicates identical distributions and 0 indicates no overlap. For discrete probability distributions, BC is defined by 
\begin{equation*}
    BC(P, Q) = \sum_{x \in \mathcal{X}} \sqrt{P(x) \cdot Q(x)}.
\end{equation*} \\ \\
\textbf{Accuracy} 
For classification accuracy, we first derive a single label, $\hat{y}$, from the model's predicted probability distribution, $P$, by taking the argmax over the label space $\mathcal{Y}$
\begin{equation*}
    \hat{y} = \text{arg}\,\max_{y \in \mathcal{Y}} P(y).
\end{equation*}
The ground truth label, $y^*$, is similarly derived from the ground truth distribution, $Q$, by selecting the label with maximum probability mass.
\begin{equation*}
    \hat{y^*} = \text{arg}\,\max_{y \in \mathcal{Y}} Q(y).
\end{equation*}
Given $N$ evaluation examples, accuracy is then defined as
\begin{equation*}
    \text{Accuracy} = \frac{1}{N} \sum_{i=1}^{N} \mathbf{1}[\hat{y}_i = y^*_i].
\end{equation*} \\ \\
\textbf{F1 Score} The F1 score is the harmonic mean of precision and recall for a given class $c$
\begin{equation*}
    \text{F1}_c = \frac{2 \cdot \text{Precision}_c \cdot \text{Recall}_c}{\text{Precision}_c + \text{Recall}_c},
\end{equation*}
where
\begin{equation*}
    \text{Precision}_c = \frac{TP_c}{TP_c + FP_c}
\end{equation*}
\begin{equation*}
    \text{Recall}_c = \frac{TP_c}{TP_c + FN_c}
\end{equation*}
and $TP_c$, $FP_c$, and $FN_c$ denote the true positives, false positives, and false negatives for class $c$, respectively. Hard predicted labels $\hat{y}$ and ground truth labels $y^*$ are derived from $Q$ and $P$ via argmax as described above. The macro-averaged F1 score averages $\text{F1}_c$ uniformly across all $|\mathcal{Y}|$ classes:
\begin{equation*}
    \text{F1}_{\text{macro}} = \frac{1}{|\mathcal{Y}|} \sum_{c \in \mathcal{Y}} \text{F1}_c.
\end{equation*} \\ \\
\textbf{Weighted F1 Score}
The weighted F1 score extends the macro-averaged F1 by accounting for class imbalance through support-weighted averaging. Specifically, the F1 score for each class $c$ is weighted by its support $N_c$, defined as the number of ground-truth instances belonging to class $c$. The weighted F1 is then computed as:
\begin{equation*}
    \text{F1}_{\text{weighted}} =
    \frac{\sum_{c \in \mathcal{Y}} N_c \cdot \text{F1}_c}{\sum_{c \in \mathcal{Y}} N_c},
\end{equation*}
where $N_c = TP_c + FN_c$ represents the total number of true instances of class $c$. This metric therefore reflects both per-class performance and the empirical class distribution, giving higher influence to more frequent classes in the dataset.

\section{Ambiguity Analysis Results for GoEmotions and MSP-Podcast}
In this section, we expand on the results in Table~\ref{tab:baseline_results} by providing an analysis of model performance at different ambiguity levels for each dataset as defined in App. \ref{app:amb_level}

%\subsection{ChaosNLI Ambiguity Level Analysis}

% \begin{table}[h]
% \centering
% \scriptsize
% \setlength{\tabcolsep}{2pt}
% \renewcommand{\arraystretch}{1.2}

% \resizebox{\columnwidth}{!}{%
% \begin{tabular}{l l c c c c c}
% \toprule
% \textbf{Method} & \textbf{Ambiguity Level} & JSD ($\downarrow$) & BC ($\uparrow$) & Acc ($\uparrow$) & F1 ($\uparrow$) & W-F1 ($\uparrow$) \\
% \midrule

% \multirow{3}{*}{ZS}
% & Low    & 0.3129 & 0.8592 & 0.6974 & 0.4915 & 0.6278 \\
% & Medium & 0.3414 & 0.8747 & 0.6187 & 0.4950 & 0.5634 \\
% & High   & 0.4351 & 0.8175 & 0.5542 & 0.4349 & 0.5009 \\
% \midrule

% \multirow{3}{*}{MLS}
% & Low    & 0.2297 & 0.9035 & 0.9079 & 0.8648 & 0.9045 \\
% & Medium & 0.4505 & 0.7801 & 0.7559 & 0.7040 & 0.7403 \\
% & High   & 0.5841 & 0.6690 & 0.5663 & 0.5174 & 0.5484 \\
% \midrule

% \multirow{3}{*}{\names(w/o Ambi-En)}
% & Low    & 0.1903 & 0.9620 & 0.9605 & 0.9390 & 0.9590 \\
% & Medium & 0.1830 & 0.9651 & 0.7860 & 0.7437 & 0.7737 \\
% & High   & 0.2022 & 0.9633 & 0.6064 & 0.5314 & 0.5770 \\
% \midrule

% \multirow{3}{*}{\names}
% & Low    & 0.1762 & 0.9666 & 0.9605 & 0.9441 & 0.9609 \\
% & Medium & 0.1817 & 0.9664 & 0.8161 & 0.8066 & 0.8139 \\
% & High   & 0.1840 & 0.9656 & 0.6506 & 0.6423 & 0.6500 \\
% \bottomrule
% \end{tabular}%
% }
% \caption{ChaosNLI evaluation metrics across different ambiguity levels.}
% \label{tab:chaos_amb_ressults}
% \end{table}

\label{app:reasoning_example}

\begin{table}[t]
\scriptsize
\centering
\caption{Observing model reasoning on a ChaosNLI example with a \textbf{High} level of ambiguity. Given the \textbf{Premise}: "In short, we all got tired of clever
analyses of what might happen; and throughout economics there was a shift in focus away from theorizing, toward data collection and careful statistical analysis" and \textbf{Hypothesis}: "We all got tired of  data collection and clever analyses of what might happen", annotators are asked to give a vote on entailment (E), neutral (N), or contradiction (C).}
\label{tab:chaos_case_high}
\setlength{\tabcolsep}{2pt}
\renewcommand{\arraystretch}{1.1}

\begin{tabularx}{\columnwidth}{|p{0.12\columnwidth}|X|p{0.05\columnwidth}|p{0.05\columnwidth}|p{0.05\columnwidth}|}
\hline
\textbf{} & \textbf{Model Reasoning} & \textbf{E} & \textbf{N} & \textbf{C} \\
\hline

Ground Truth & & .45  & .17 & .38 \\

\hline

Baseline & 
1. **Premise Analysis**:
   - The premise states that there was a shift in focus away from theorizing and clever analyses towards data collection and careful statistical analysis.
   - It implies that people got tired of clever analyses and moved towards data collection.

2. **Hypothesis Analysis**:
   - The hypothesis states that people got tired of data collection and clever analyses of what might happen.
   - This is the opposite of what the premise suggests.

3. **Relationship Between Premise and Hypothesis**:
   - The premise and hypothesis are in direct contradiction. The premise suggests a shift towards data collection, while the hypothesis suggests a shift away from data collection.

4. **Multiple Valid Interpretations**:
   - There are no multiple valid interpretations here. The premise and hypothesis are clearly in conflict.

5. **Probability Assignment**:
   - Since the premise and hypothesis are in direct contradiction, the probability of entailment is 0.
   - The probability of neutrality is also 0 because the relationship is not neutral.
   - The probability of contradiction is 1 because the hypothesis directly contradicts the premise.

& .00 & .00 & 1.00 \\

\hline

$R_1$ &
The relationship between the premise and the hypothesis is somewhat complex. The premise states that there was a shift in focus away from theorizing and towards data collection and statistical analysis, which implies that the focus on clever analyses of what might happen was lessened. However, the hypothesis suggests that we got tired of data collection and clever analyses of what might happen, which is somewhat contradictory to the premise.

There are multiple valid interpretations:
1. The premise and hypothesis could be seen as contradictory because the premise indicates a shift away from clever analyses, while the hypothesis suggests tiredness with both data collection and clever analyses.
2. Alternatively, the premise could be interpreted as a shift in focus, which might imply that the tiredness with clever analyses was a result of the shift towards data collection and statistical analysis.

Given these interpretations, the probability of entailment is relatively low, as the hypothesis suggests a contradiction to the premise. The probability of neutrality is moderate, as the premise and hypothesis could be seen as having different focuses. The probability of contradiction is high, as the hypothesis directly contradicts the premise.

& .20 & .50 & .30 \\

\hline
\end{tabularx}
\end{table}

\begin{table}[t]
\scriptsize
\centering
\caption{Observing model reasoning on a ChaosNLI example with a \textbf{Medium} level of ambiguity. Given the \textbf{Premise}: A child in a red jacket, waist deep in a pit on the beach and \textbf{Hypothesis}: A child is building a sandcastle on the beach, annotators are asked to give a vote on entailment (E), neutral (N), or contradiction (C).}
\label{tab:chaos_case_med}
\setlength{\tabcolsep}{2pt}
\renewcommand{\arraystretch}{1.1}

\begin{tabularx}{\columnwidth}{|p{0.12\columnwidth}|X|p{0.05\columnwidth}|p{0.05\columnwidth}|p{0.05\columnwidth}|}
\hline
\textbf{} & \textbf{Model Reasoning} & \textbf{E} & \textbf{N} & \textbf{C} \\
\hline

Ground Truth & & .00  & .77 & .23 \\

\hline

Baseline & 
1. **Premise Analysis**:
   - The premise describes a child in a red jacket, waist deep in a pit on the beach. This suggests the child is engaged in some form of activity in the sand, but it does not specify what that activity is.

2. **Hypothesis Analysis**:
   - The hypothesis states that a child is building a sandcastle on the beach. This is a specific activity that involves creating a sandcastle.

3. **Relationship Between Premise and Hypothesis**:
   - The premise does not provide enough information to confirm or deny the hypothesis. The child could be building a sandcastle, but they could also be doing other activities like digging, playing, or just standing in the sand.

4. **Multiple Valid Interpretations**:
   - There are multiple valid interpretations of the premise. The child could be building a sandcastle, but they could also be doing other activities. This lack of specificity in the premise means that the hypothesis is neither entailed nor contradicted by the premise.

5. **Probability Assignment**:
   - Since the premise does not provide enough information to confirm or deny the hypothesis, the relationship between the two statements is neutral. The probability of entailment is low because the premise does not logically follow the hypothesis. The probability of contradiction is also low because the premise does not conflict with the hypothesis. The probability of neutrality is high because the premise does not provide enough information to determine the relationship.

& .05 & .90 & 0.05 \\

\hline

$R_1$ &
The relationship between the premise and the hypothesis is neutral. The premise describes a child in a red jacket, waist deep in a pit on the beach, but it does not provide enough information to determine if the child is building a sandcastle. The child could be doing various activities such as playing, digging, or even just standing in the pit. There are multiple valid interpretations of the premise, which makes it difficult to assign a high probability to either entailment or contradiction. Therefore, the probability of entailment is low, the probability of neutrality is high, and the probability of contradiction is also low.

& .10 & .80 & .10 \\

\hline
\end{tabularx}
\end{table}

\begin{table}[t]
\scriptsize
\centering
\caption{Observing model reasoning on a ChaosNLI example with a \textbf{Low} level of ambiguity. Given the \textbf{Premise}: Linda Tripp was indicted for illegally taping telephone conversations with Monica Lewinsky and \textbf{Hypothesis}: Monica Lewinsky was illegally taped by Linda Tripp, annotators are asked to give a vote on entailment (E), neutral (N), or contradiction (C).}
\label{tab:chaos_case_low}
\setlength{\tabcolsep}{2pt}
\renewcommand{\arraystretch}{1.1}

\begin{tabularx}{\columnwidth}{|p{0.12\columnwidth}|X|p{0.05\columnwidth}|p{0.05\columnwidth}|p{0.05\columnwidth}|}
\hline
\textbf{} & \textbf{Model Reasoning} & \textbf{E} & \textbf{N} & \textbf{C} \\
\hline

Ground Truth & & .95  & .03 & .02 \\

\hline

Baseline & 
1. **Premise Analysis**:
   - The premise states that "Linda Tripp was indicted for illegally taping telephone conversations with Monica Lewinsky."
   - This implies that Linda Tripp was the one who performed the illegal act of taping.

2. **Hypothesis Analysis**:
   - The hypothesis states that "Monica Lewinsky was illegally taped by Linda Tripp."
   - This suggests that Monica Lewinsky was the one who was taped by Linda Tripp.

3. **Relationship Between Premise and Hypothesis**:
   - The premise and hypothesis are describing the same event but from different perspectives.
   - The premise focuses on Linda Tripp as the tapper, while the hypothesis focuses on Monica Lewinsky as the taped person.
   - Since the premise explicitly states that Linda Tripp was the one who taped, the hypothesis directly contradicts this by suggesting that Monica Lewinsky was the one who was taped.

4. **Multiple Valid Interpretations**:
   - There are no multiple valid interpretations here. The premise and hypothesis are in direct conflict.

5. **Probability Assignment**:
   - Given the direct contradiction between the premise and the hypothesis, the probability of the hypothesis being true given the premise is 0.
   - The probability of the hypothesis being neutral or entailment is also 0 because the hypothesis directly contradicts the premise.
& .00 & .00 & 1.00 \\

\hline

$R_1$ &
The relationship between the premise and the hypothesis is that the hypothesis logically follows from the premise. The premise states that Linda Tripp was indicted for illegally taping telephone conversations with Monica Lewinsky, which directly implies that Monica Lewinsky was the subject of the illegal taping by Linda Tripp. \textbf{There is no ambiguity in this relationship}, and the hypothesis is a direct consequence of the premise.

Given this clear relationship, the probability of entailment is high. The hypothesis is neutral in terms of additional information not provided in the premise, and there is no contradiction since the hypothesis is a logical outcome of the premise.
& .85 & .10 & .05 \\

\hline
\end{tabularx}
\end{table}

%\subsection{MSP Podcast Ambiguity Level Analysis}

\begin{figure*}[!tb]
\centering
\includegraphics[width=\linewidth]{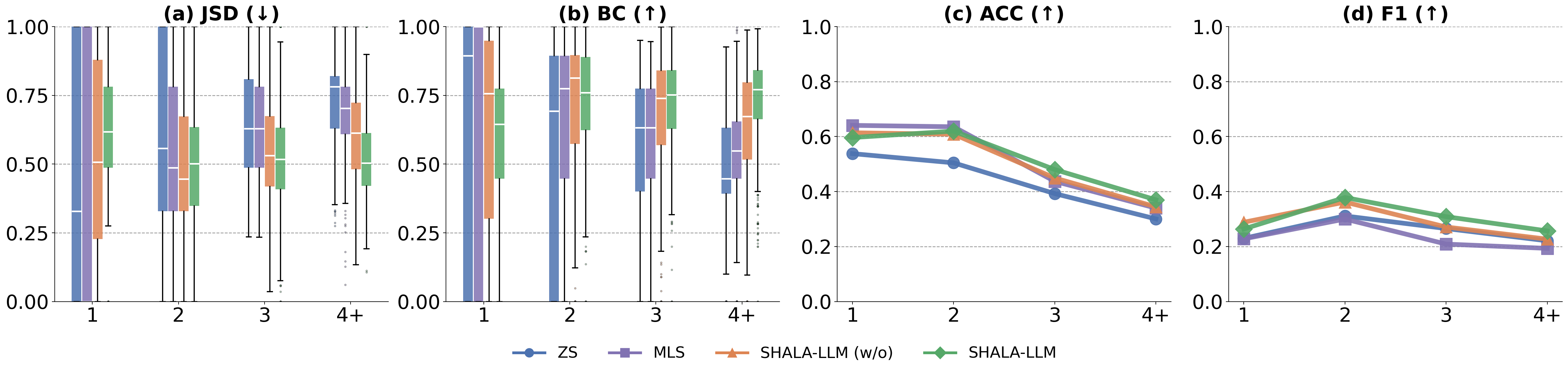}
\caption{Performance comparison across different ambiguity levels on the \textbf{MSP Podcast} dataset. x-axis denotes the number of active labels in the ground truth distribution which reflects ambiguity levels from low (left) to high (right) explained in App. \ref{app:msp_ambi}}
\label{fig:msp_ambi_level}
\end{figure*}

%\subsection{GoEmotions Ambiguity Level Analysis}

\begin{figure*}[!tb]
\centering
\includegraphics[width=\linewidth]{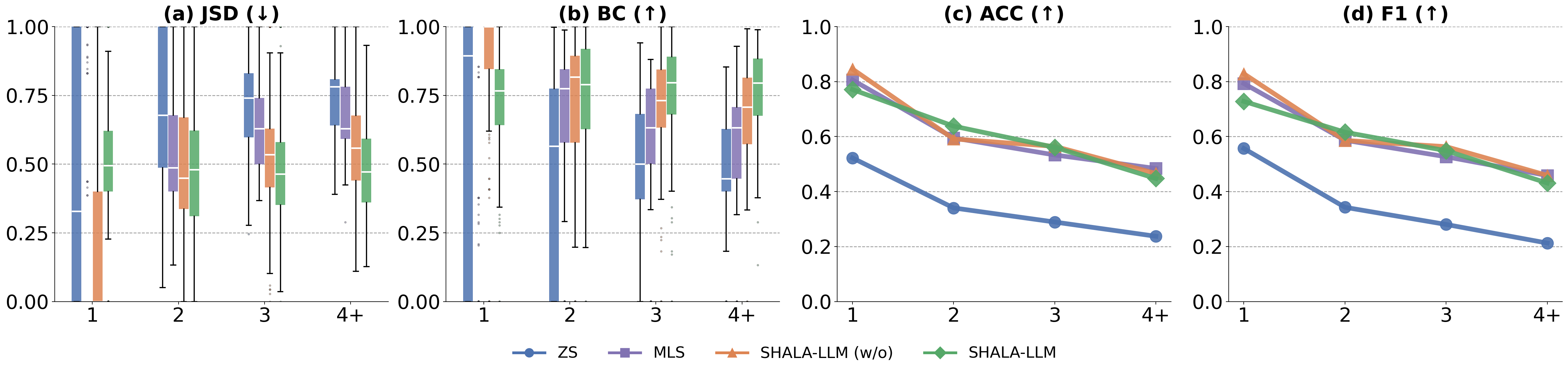}
\caption{Performance comparison across different ambiguity levels on the \textbf{GoEmotions} dataset. x-axis denotes the number of active labels in the ground truth distribution which reflects ambiguity levels from low (left) to high (right) explained in App. \ref{app:goemo_ambi}}
\label{fig:goemo_ambi_level}
\end{figure*}

We provide an in-depth analysis of model performance across different ambiguity levels on GoEmotions and MSP-Podcast in Fig.~\ref{fig:goemo_ambi_level} and \ref{fig:msp_ambi_level}. Both \names~ and its variant, \names~ (w/o Ambi-En), consistently outperform the ZS and MLS baselines across all ambiguity levels and evaluation metrics, particularly on the distributional metrics JSD and BC. Importantly, as ambiguity increases, \names~ exhibits the smallest performance degradation, demonstrating the robustness of ambiguity-aware training under highly ambiguous conditions. Although \names and \names~ (w/o Ambi-En) do not show substantial gains over MLS on conventional classification metrics compared with the improvements over ZS, this is likely because MLS is primarily optimized for dominant single-label classification. Overall, \names~ demonstrates strong capability in both capturing annotator distributions, as evidenced by JSD and BC, and maintaining competitive performance on dominant-label classification, further supporting the discussion in Section~\ref{sec:ambi_level}.

\section{Per-Class Analysis Results}
%\jw{after adding the figure, we need to add a few sentence expplanations too}
In this section, we expand on the results in Table~\ref{tab:baseline_results} by providing an analysis of model performance at the per-class level.

% \subsection{Detailed ChaosNLI Results}

\subsection{MSP Podcast Per-Class Analysis}

%Fig.~\ref{fig:msp_per_class} shows the per class analysis for different models. It is observed that on distributional metric, \names\ and it's variation \names\ (w/o) consistently outperform the baseline ZS and MLS, while maintaining relatively well performance on converntional metric. Importantly, improvements concentrated in happy class, mostly likely due to class imbalance (happy class is more frequent)
%suprisingly the contempt class does poorly in mls and shalla-llm (w/o) but acc and f1 drastically increases with shala-llm. This demonstract that xxxx

\begin{figure*}[!tb]
\centering
\includegraphics[width=\linewidth]{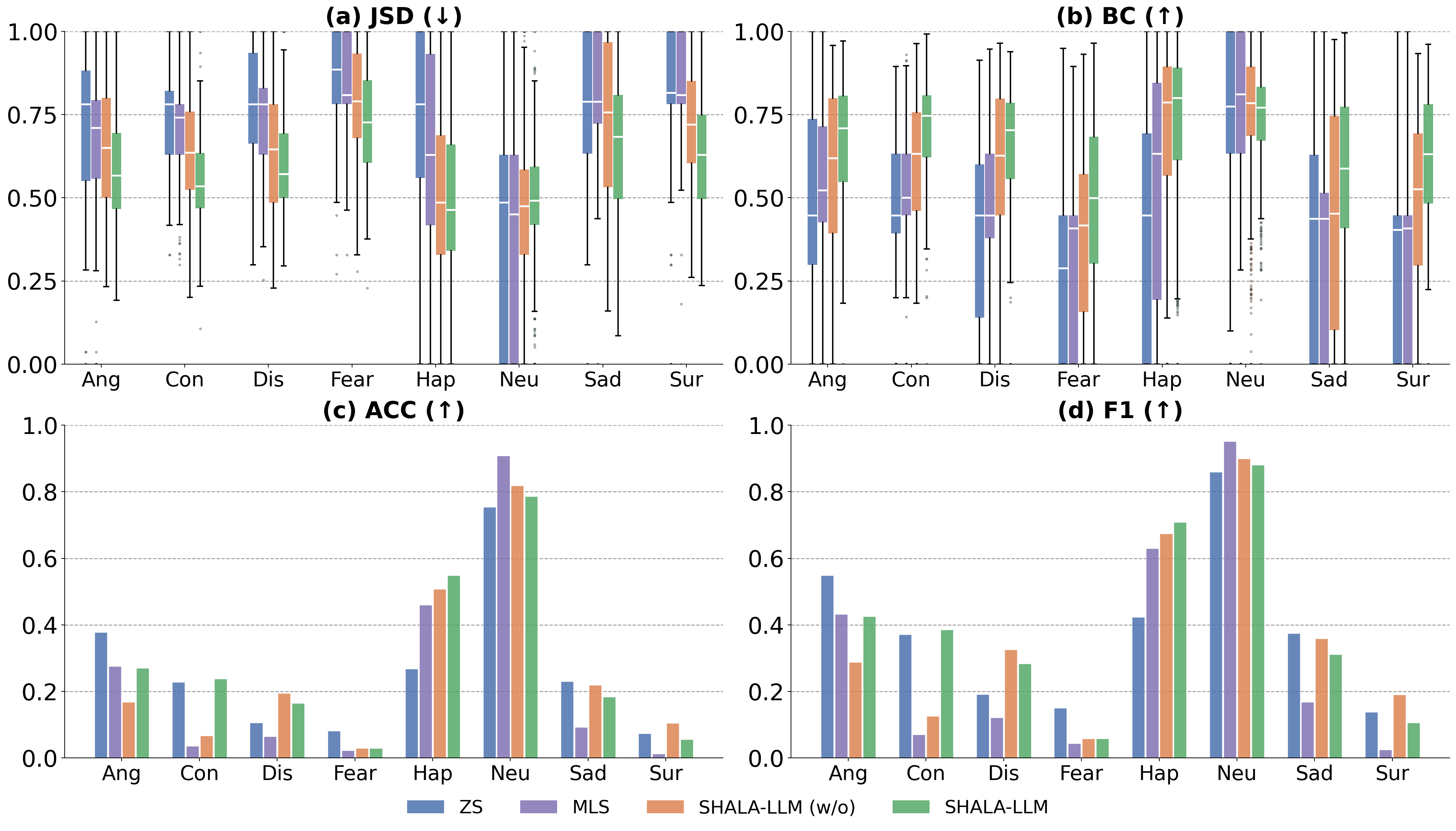}
\caption{Performance comparison across different emotion classes on the \textbf{MSP Podcast} dataset.}
\label{fig:msp_per_class}
\end{figure*}

Fig.~\ref{fig:msp_per_class} presents the per-class analysis across different models. It is observed that, on distributional evaluation metrics, both \names\ and its ablation variant \names\ (w/o) consistently outperform the baseline ZS and MLS models, while still maintaining relatively strong performance on conventional classification metrics. Importantly, the largest improvements are concentrated in the \textit{Happy} class, likely due to class imbalance, as this class contains substantially more training samples. Interestingly, the \textit{Contempt} class exhibits relatively poor performance under MLS and \names\ (w/o), whereas the full \names\ framework substantially improves both ACC and F1. This suggests that ambiguity-enhanced optimization helps the model better capture highly ambiguous or underrepresented emotional categories by amplifying informative disagreement structures during learning.

\subsection{GoEmotions Per-Class Analysis}

%Figure~\ref{fig:goemo_per_class} depicts the per-class analysis on GoEmotion dataset. It is observed that for distributional evaluation, both \names\ and \names\ (w/o) showed strong performance compared to ZS and MLS and generally high performance with classiifcation metric. Particularly on high ambiguous class such as \textit{Surprise} which the dataset collector designed for ambiguity, although this is not obviously seen in classification metric due to it's ambiguous property.

\begin{figure*}[!tb]
\centering
\includegraphics[width=\linewidth]{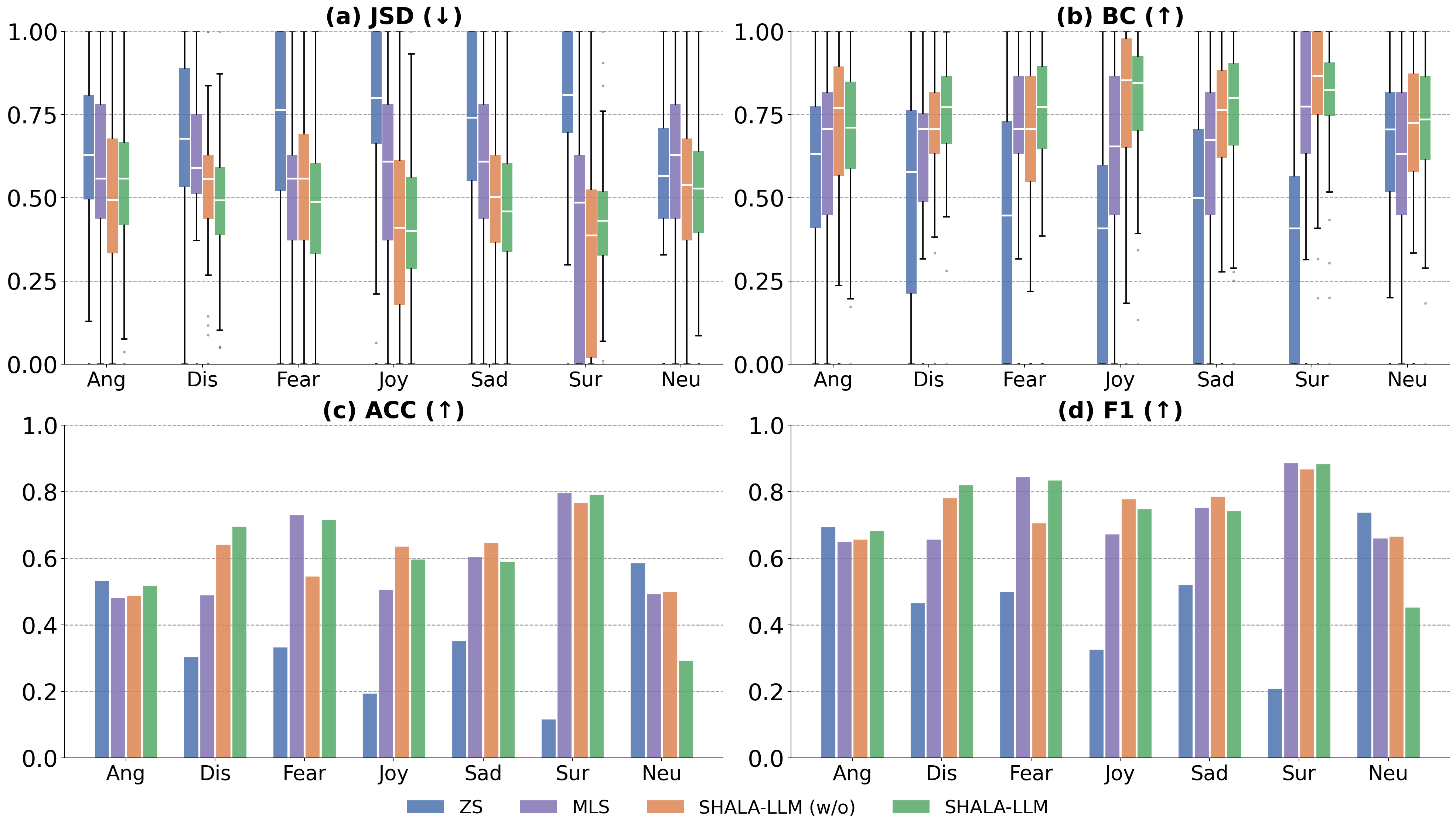}
\caption{Performance comparison across different emotion classes on the \textbf{GoEmotions} dataset.}
\label{fig:goemo_per_class}
\end{figure*}

Fig.~\ref{fig:goemo_per_class} presents the per-class analysis on the GoEmotions dataset. Both \names\ and \names\ (w/o Ambi-En) demonstrate strong performance compared with the ZS and MLS baselines on distributional evaluation metrics, while also maintaining generally competitive classification performance. Notably, improvements are particularly evident for highly ambiguous categories such as \textit{Surprise}, which was intentionally designed by the dataset creators to exhibit ambiguity. Although these improvements are less apparent on conventional classification metrics due to the inherently ambiguous nature of the category, the distributional metrics reveal that \names\ better captures the underlying annotator disagreement.

\section{Model Reasoning Under Different Ambiguity Levels}

%\jw{Ahsley, include the prompt here.}

We further include model reasoning outputs for representative low-, medium-, and high-ambiguity examples in Tables~\ref{tab:chaos_case_high} to \ref{tab:chaos_case_low}. For low- and medium-ambiguity cases, we do not observe substantial differences between \names\ and the baseline. However, for highly ambiguous cases, \names\ demonstrates a stronger capability to reason over multiple valid interpretations, as discussed in detail in Section~\ref{sec:model_reasoning} of the main paper.

We include the following prompt during inference to encourage the model to explain its reasoning.

\begin{mdframed}
Before producing your final answer, explain your reasoning:
    \begin{itemize}
        \item What is the relationship between the premise and hypothesis?"
        \item Are there multiple valid interpretations? If so, describe them.
        \item How does your reasoning inform the probability you assign to each label? 
    \end{itemize}
Then output EXACTLY this JSON on its own line: \{\{"entailment": float, "neutral": float, "contradiction": float\}\}
\end{mdframed}

\section{Use of AI Assistants}\label{app:ai_use}
AI assistants were used for language refinement, editing support during manuscript preparation and debugging for codes. All technical content, experimental design, implementation, and scientific claims were developed and verified by the authors.

\end{document}